\documentclass{article}
\usepackage{geometry}
\usepackage{xcolor}
\usepackage[utf8]{inputenc}
\usepackage[T1]{fontenc}
\usepackage[hyphens]{url}
\usepackage[colorlinks,pdftex,pdfpagelabels,bookmarks,hyperindex,hyperfigures]{hyperref}
\hypersetup{
  pdftitle={},
  pdfauthor={Reza Sameni},
    pdfnewwindow=true,      % links in new window
    colorlinks=true,       % false: boxed links; true: colored links
    linkcolor=blue,          % color of internal links
    citecolor=magenta,        % color of links to bibliography
    filecolor=cyan,      % color of file links
    urlcolor=teal%magenta           % color of external links		
}

\usepackage{booktabs}
\usepackage{amsfonts}
\usepackage{color}

\usepackage{amsmath}
\usepackage{amssymb}
\usepackage{graphicx}
\usepackage{subcaption}
\usepackage{algorithm}
\usepackage{algorithmic}
\usepackage{balance}

\begin{document}
% \maketitle

% \bstctlcite{IEEEexample:BSTcontrol}

% \history{Date of publication xxxx 00, 0000, date of current version xxxx 00, 0000.}
% \doi{10.1109/ACCESS.2017.DOI}

% \title{Generation of synthetic paper electrocardiogram records for deep learning based digitization}
% \title{A Deep Learning-Based Approach for Signal and Data Recovery from Scanned ECG Records}

% \title{A Synthetic Electrocardiogram (ECG) Image Generation Toolbox to Facilitate Deep Learning-Based Scanned ECG Digitization}
\title{ECG-Image-Kit: A Synthetic Image Generation Toolbox to Facilitate Deep Learning-Based Electrocardiogram Digitization}

%//////////////////////////////////////////
\author{Kshama~Kodthalu~Shivashankara, Deepanshi, Afagh~Mehri~Shervedani,\\ Gari~D.~Clifford, Matthew~A.~Reyna, and Reza~Sameni
\thanks{K.~Kodthalu~Shivashankara is with the School of Electrical and Computer Engineering, Georgia Institute of Technology, Atlanta, GA, USA. A. Mehri~Shervedani is with the Electrical and Computer Engineering Department, University of Illinois Chicago, Chicago, IL. M.A.~Reyna, G.D.~Clifford and R.~Sameni are with the Department of Biomedical Informatics, Emory University, Atlanta, GA, USA. G.D.~Clifford is also with the Department of Biomedical Engineering, Georgia Institute of Technology, GA, USA. Corresponding author: R.~Sameni (\href{mailto:rsameni@dbmi.emory.edu}{rsameni@dbmi.emory.edu})
}}

% \author{\uppercase{Kshama~Kodthalu~Shivashankara}\authorrefmark{1}, 
%          \uppercase{Deepanshi}\authorrefmark{2},
%         \uppercase{Afagh~Mehri~Shervedani}\authorrefmark{3},
%         \uppercase{Matthew~A.~Reyna}\authorrefmark{2},
%         \uppercase{Gari~D.~Clifford}\authorrefmark{2, 4}
%         \IEEEmembership{Fellow,~IEEE},
%         \uppercase{Reza~Sameni}\authorrefmark{3}
%         \IEEEmembership{Senior Member,~IEEE}}
        
% \address[1]{School of Electrical and Computer Engineering, Georgia Institute of Technology, Atlanta, GA, USA.}
% \address[2]{Department of Biomedical Informatics, Emory University, Atlanta, GA, USA.}
% \address[3]{Electrical and Computer Engineering Department, University
% of Illinois Chicago, Chicago, IL, USA.}
% \address[4]{Biomedical Engineering Department, Georgia Institute of Technology, Atlanta, GA, USA.}
%\tfootnote{R.~Sameni is supported by ...}

\markboth % \headeretal
{K.~K.~Shivashankara: A Synthetic ECG Image Generation Toolbox for ECG Digitization}
{K.~K.~Shivashankara: A Synthetic ECG Image Generation Toolbox for ECG Digitization}

% \corresp{Corresponding author: Reza~Sameni (e-mail: \href{mailto:rsameni@dbmi.emory.edu}{rsameni@dbmi.emory.edu}).}

\date{}

% \begin{document}
\maketitle
\begin{abstract}
Cardiovascular diseases are a major cause of mortality globally, and electrocardiograms (ECGs) are crucial for diagnosing them. Traditionally, ECGs are stored in printed formats. However, these printouts, even when scanned, are incompatible with advanced ECG diagnosis software that require time-series data. Digitizing ECG images is vital for training machine learning models in ECG diagnosis, leveraging the extensive global archives collected over decades. Deep learning models for image processing are promising in this regard, although the lack of clinical ECG archives with reference time-series data is challenging. Data augmentation techniques using realistic generative data models provide a solution.

We introduce \textit{ECG-Image-Kit}, an open-source toolbox for generating synthetic multi-lead ECG images with realistic artifacts from time-series data, aimed at automating the conversion of scanned ECG images to ECG data points. The tool synthesizes ECG images from real time-series data, applying distortions like text artifacts, wrinkles, and creases on a standard ECG paper background.

As a case study, we used ECG-Image-Kit to create a dataset of 21,801 ECG images from the PhysioNet QT database. We developed and trained a combination of a traditional computer vision and deep neural network model on this dataset to convert synthetic images into time-series data for evaluation. We assessed digitization quality by calculating the signal-to-noise ratio (SNR) and compared clinical parameters like QRS width, RR, and QT intervals recovered from this pipeline, with the ground truth extracted from ECG time-series. The results show that this deep learning pipeline accurately digitizes paper ECGs, maintaining clinical parameters, and highlights a generative approach to digitization. This toolbox currently supports data augmentation for the 2024 PhysioNet Challenge, focusing on digitizing and classifying paper ECG images.
\end{abstract}

% \begin{IEEEkeywords}
% Electrocardiogram (ECG), ECG digitization, deep learning, synthetic data, denoising CNN, data augmentation
% \end{IEEEkeywords}

% \titlepgskip=-15pt

\maketitle

\section{Introduction}
Cardiovascular diseases (CVDs) are the primary cause of mortality globally among adults aged 37 to 70 years \cite{DAGENAIS2020785}, and the electrocardiogram (ECG) is the most accessible and widely used tool for CVD diagnosis. Clinical ECG is most accurately studied through standard 12-lead recordings. Every day, clinicians conduct millions of ECGs, and wearable and personal devices generate millions more. There are billions of digital diagnostic ECGs globally and billions more in conventional formats such as microfilms, printed papers and scanned images. Although this legacy contains invaluable information on prevalent and rare CVDs and their evolution across generations and geography, we are not currently ``learning'' from ECG archives. Due to natural deterioration, lack of funding and a transition to digital ECGs, non-digital ECG archives worldwide will soon be destroyed, before we can learn from them. This will be an irreversible loss for CVD research, since the ECG is the only biological signal that has been recorded for over a century without significant changes in its acquisition protocol, especially for low and low and middle-income countries (LMICs), where paper ECGs are still more common. Importantly, the ECG has been acquired globally for decades, without significant changes in its acquisition protocol. As a result, ECG data is abundant, and beyond human experts' capacity in prescreening these data. Machine learning (ML) algorithms can help automate the process of ECG-based diagnostic decision-making. However, paper ECGs or scanned ECG images are not compatible with state-of-the-art ML algorithms that are trained and tested on ECG time-series. Although non-digital ECGs can be scanned and archived as images, there is little incentive to do so; because currently ECG images are not automatically searchable for anomalies, and are incompatible with annotation software that analyze ECG time-series \cite{siontis2021artificial}. For example, ECG biomarkers, such as RR, PR, QRS, QT intervals, waveforms and rhythms are easily accessible from ECG time-series and could help improve the performance of existing ML models. Digitization of the existing paper ECG archives is therefore essential.

Furthermore, access to paper ECGs are limited primarily by privacy concerns, as they contain personal and sensitive information. The Health Insurance Portability and Accountability Act (HIPAA) mandates safeguards for protecting the privacy of health information, often necessitating patient consent for data usage, thereby reducing the availability of open-source datasets for training ML models \cite{annas2003hipaa}. Consequently, there is a growing interest in modeling ECGs synthetically in a privacy-preserving way. Our work focuses on producing synthetic paper-like ECG images from ground truth time-series data, facilitating the training of ML and deep learning digitization solutions within privacy constraints. We demonstrate the toolbox's utility by developing a deep learning ECG digitization model and extracting clinical parameters to establish the accuracy of the model for clinical applications. Our developed tool for synthetic ECG image generation and digitization has been implemented in Python and is available online in the ECG-Image-kit toolbox: \url{https://github.com/alphanumericslab/ecg-image-kit}. The toolbox has also been used for the 2024 PhysioNet Challenge on the digitization and classification of ECG images \cite{ChallengesWebsite,2024ChallengeWebsite} 

\section{Related work}
The growing interest in generating realistic synthetic data and \textit{digital twins}, driven by privacy preservation and HIPAA mandates, has led to innovations in synthetic medical data generation, especially in Electronic Health Records (EHR). John et al.\ utilized a federated generative adversarial network (GAN) for EHR generation~\cite{weldon2021generation}; Choi et al.\ demonstrated EHR generation through medGAN~\cite{choi2017generating}; and EHR-Safe proposed a sequential encoder-decoder architecture with GANs~\cite{Yoon2023}.

Accurate ECG digitization is crucial for advancing cardiology patient treatment and research. Automated analysis of digitized ECGs facilitates early cardiovascular disease diagnosis. 

In the ECG context, various digitization methods have been proposed; including the grayscale thresholding and contour-based digitization method by Ravichandran et al.\ \cite{6527311}, color segmentation and median filtering for noise removal by Garg et al. \cite{garg2012ecg}, and the combination of optical character recognition with image processing techniques for digitization and artifact removal \cite{ganesh2021combining}, \cite{baydoun2019high}. However, classical image processing methods, sensitive to input quality and environmental artifacts, often struggle with low-quality, distorted paper ECG records.

Recent advancements in deep learning have also been applied to the digitization of paper ECG records. Mishra et al.\ used deep learning to determine the binarization threshold for grid removal \cite{mishra2021ecg}. Li et al.\ approached digitization as a segmentation problem, employing the U-Net architecture for digitizing noisy ECG scans \cite{li2020deep}. These methods are effective but face limitations due to a lack of diverse noise artifacts, such as handwritten text and environmental reflections, in the training datasets. Additionally, there is a scarcity of paper ECG records with ground truth time-series data representing noisy records.

Our research addresses these gaps by utilizing deep learning for paper ECG digitization and enriching the training dataset with synthetic images from realistic generative models. Numerous studies have illustrated synthetic time-series ECG generation, from dynamical models simulating adult and fetal ECGs \cite{mcsharry2003dynamical,sameni2007multichannel} to artificial vector models for abnormal rhythms \cite{clifford2010artificial} and signal decomposition for morphological modeling \cite{Roonizi2013}. Recent advances include GAN-based synthetic ECG generation, with applications ranging from simulating various cardiac conditions \cite{zhu2019electrocardiogram, zhang2021synthesis, wulan2020generating} to creating comprehensive repositories of \textit{DeepFake} ECGs \cite{thambawita2021deepfake}. These developments underscore the feasibility of generating synthetic time-series ECG using generative models. Our approach uniquely combines privacy-preserving synthetic ECG image generation with standard paper ECG backgrounds and sequential artifact addition, creating synthetic paper-like ECG images with underlying real ECG time-series data.

\section{Synthetic ECG image generation pipeline}
\begin{figure*}[tb]
     \centering
         \includegraphics[width=\columnwidth]{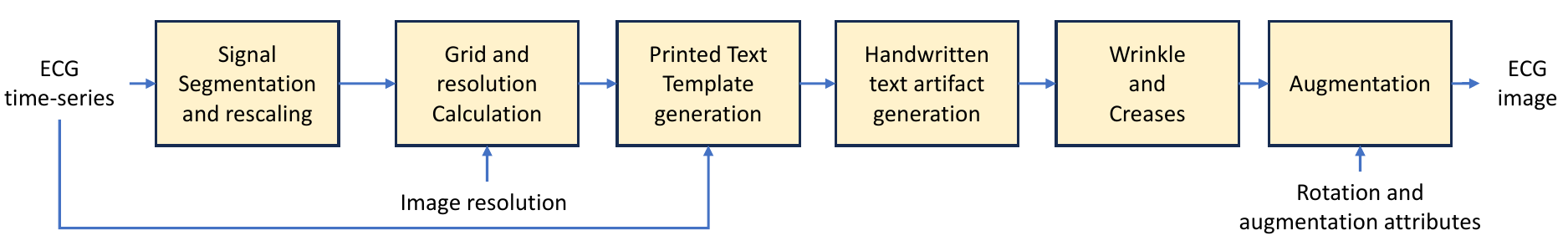}
         \caption{Proposed pipeline for generating synthetic ECG images}
         \label{fig:methodology}
\end{figure*}

The proposed methodology involves multiple stages. First, a synthetic paper ECG dataset is created. This process includes adding distortions step-by-step to time-series data plotted on standard ECG grids, as shown in Fig.~\ref{fig:methodology} and detailed in the sequel.

\subsection{Background}

\subsubsection{The standard ECG paper format}
Standard surface ECG acquisition records heart activity using a 12-lead system with ten electrodes on the body, including the limbs and chest. This setup includes three limb leads (I, II, III), three augmented limb leads (aVR, aVL, aVF), and six precordial leads (V1--V6). These leads, despite some geometrical redundancy, offer comprehensive cardiac perspectives, crucial for diagnosing arrhythmias, myocardial infarction, and other heart conditions \cite{MalmivuoPlonsey1995}. The limb leads provide frontal plane views, while precordial leads assess the heart's horizontal plane. Recent advancements include reduced lead ECG systems, using fewer electrodes with computational methods, including machine and deep learning, to reconstruct a complete 12-lead ECG \cite{Reyna2022, PerezAlday2020, Whyte2023, Dwivedi2023}.

\begin{figure}[tb]
 \centering
\includegraphics[width=0.3\columnwidth]{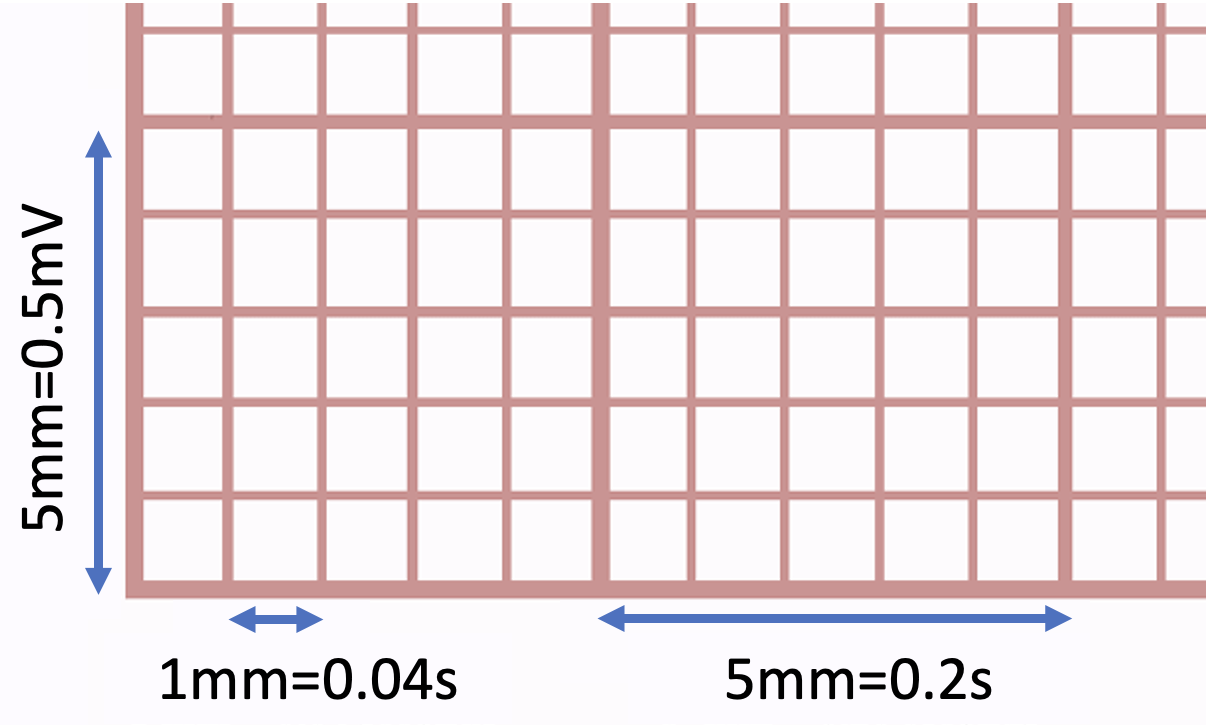}
     \caption{The standard grid on printed ECG papers/images}
     \label{fig:ECG_grid}
\end{figure}

Conventionally, analog and digital ECG machines printed ECGs on so-called thermal paper at a horizontal speed of 25\,mm/sec and a vertical scale of 0.1\,mV per 10\,mm. Modern ECG machines, whether printing hard copies or generating PDF images, use the same convention. The standard paper ECG features two grids: a coarse grid of 5\,mm$\times$5\,mm corresponding to 0.5\,mV in the vertical (amplitude) and 0.2\,s in the horizontal (time) directions, and a fine grid of 1\,mm$\times$1\,mm corresponding to 0.1\,mV and 40\,ms in vertical and horizontal directions, as shown in Fig.~\ref{fig:ECG_grid}. Historically, a calibration pulse of 1\,mV amplitude and 0.2\,s width is also printed on most paper ECGs \cite{luthra2019ecg}. 

While most paper ECG grids are red-pink in color, there is no widely accepted standard for ECG paper color. Modern digital ECGs are typically generated as PDF files generated for A4 or US Letter-sized papers. Standard paper ECGs usually display all 12 leads in 2.5\,s segments over four rows. Additionally, leads II, V1, V2, or V5 are often plotted as a continuous 10\,s strip at the bottom, for rhythm analysis. Older ECG machines swept the 2.5\,s segments across different leads asynchronously. Therefore, the 2.5\,s segments of the different leads did not correspond to the same time frame. This is an important point of caution for ECG digitization algorithms, as they would not be able to benefit from the synchrony of the channel segments to improve the extracted ECG time-series through multichannel post-processing.

Although the majority of paper ECG records follow the 12-lead representation (3$\times$4 segments + 1 strip), there are printed ECGs that do not adhere to this format. To account for the variability across real paper ECG records, ECG-Image-Kit enables users to adjust the lead format.

\subsubsection{ECG image vs time-series temporal and amplitude resolutions}
\label{sec:ecg_image_to_signal_resolution}
The ECG digitization process involves several key parameters: the length of the ECG segment $T$ (in seconds), the time-series sampling frequency $f_s$, the scanned image resolution in dots-per-inch (DPI, denoted as $D$), and the amplitude resolution, which in digital ECG devices is related to the analog-to-digital converter (ADC) resolution and the analog input dynamic range. Understanding these parameters is crucial for aligning the digitized ECG with the original time-series.

Printing and rescanning an ECG involves interpolation and resampling. In analog devices or printers, this process converts discrete time samples into a continuous waveform on paper. The original sampling frequency $f_s$ and the ADC resolution become irrelevant once printed, as the signal reverts to a continuous form. Upon scanning, the ECG is quantized and resampled as a two-dimensional image at a resolution of $D$ DPI. Each 1-inch square of the ECG is digitized into a $D \times D$ array, each pixel represented in $B$ bits. Typically, $B = 8$, yielding 24 bits or 3 bytes per pixel.

When a standard ECG, printed on A4 or Letter-size paper, is scanned, each 1-inch segment corresponds to $D$ pixels. Each coarse ECG square (0.5\,mV amplitude, 200\,ms time) maps to a pixel square of $(\frac{5 \times D}{25.4})\times(\frac{5 \times D}{25.4})$. Therefore, the amplitude resolution of the scanned ECG is $dv = \frac{2.54}{D}$ millivolts, and the temporal resolution is $dt = \frac{1.016}{D}$ seconds, resulting in an image-based sampling frequency of
\begin{equation}
\tilde{f}_s = \frac{D}{1.016}\,\text{Hz}
\label{eq:image_fs}
\end{equation}
As we see, this frequency is independent of the original $f_s$, and increasing $D$ yields smoother waveforms but does not add information beyond $f_s/2$ (which is bounded by the anti-aliasing filter of the original ECG device's analog front-end). From \eqref{eq:image_fs}, we may conclude that in order to preserve the typical ECG spectrum that is dominantly below 100\,Hz, a resolution of at least 200 DPI is recommended for ECG scanning and digitization (assuming that the image is full-screen, utilizing all the image DPI for the ECG image).

The accurate calculation of ECG grid size from the image DPI and paper size is reliable only when using a standard full-paper size scanner. However, for ECG images captured by cameras, smartphones, screenshots, or altered through cropping, resizing or compression, the equivalency of 1 inch on the actual paper to the captured image DPI may not be accurate. Therefore, image file metadata DPI can be unreliable for recovering pixel-wise time and amplitude resolutions. In this case, ECG digitization algorithms may employ techniques that directly analyze the ECG grid sizes from the image, using image processing methods that for instance utilize pixel marginal distributions or spectral methods to detect the regular ECG grid patterns. ECG-Image-Kit offers multiple algorithms for these purposes.

In the final stage, to recover the ECG time-series at its original sampling frequency, the digitized signal can be resampled from $\tilde{f}_s$ back to $f_s$. This enables alignment and comparison between the original and reconstruction time-series. This step is also crucial for maintaining the integrity of the ECG data and ECG-based measurements, including RR-intervals and QT-intervals.

\subsubsection{The ECG time-series dataset used for model training}
The synthetic paper ECG generation pipeline requires time-series data as the ground truth. For this purpose, we used the PTB-XL clinical ECG dataset \cite{wagner2020ptb,Goldberger2000}. The dataset contains 21,801 clinical 12-lead ECGs from 18,869 patients, each of a 10-second duration. It also includes extensive metadata and statistics on signal properties and demographics such as age, sex, height, and weight. Each record provides the standard set of 12 ECG leads (I, II, III, aVR, aVL, aVF, V1, V2, V3, V4, V5, and V6) \cite{wagner2020ptb}. In addition to the PTB-XL dataset, we used other 12-lead clinical ECG datasets such as the CPSC and CPSC-Extra Databases \cite{liu2018open}, the INCART Database \cite{tihonenko2008st}, the Georgia 12-lead ECG database (G12EC), and the PTB database \cite{bousseljot1995nutzung}. These datasets were used as part of the 2021 PhysioNet Challenges on multilead ECG annotation \cite{Reyna2022}. Segments of these data were extracted as short 10-second time-series arrays from the entire data, to construct standard 12-lead ECGs. Real ECG time-series recorded in real environments can be contaminated by various types of measurement noises, such as baseline artifacts, powerline interference, motion artifacts, muscle noise, and additive device noises \cite{clifford2006ecg,gmoody}. We added a combination of these noises to the ECG time-series (prior to conversion to images) at different levels of SNR, using the PhysioNet noise stress-test dataset \cite{gmoody}, and the synthetic noise generator from the open-source electrophysiological toolbox (OSET) \cite{sameni2007multichannel, OSET3.14}.

\subsection{Printed text artifacts}
\begin{figure}[tb]
     \centering
     \begin{subfigure}[b]{0.49\textwidth}
         \centering
         \includegraphics[width=\textwidth]{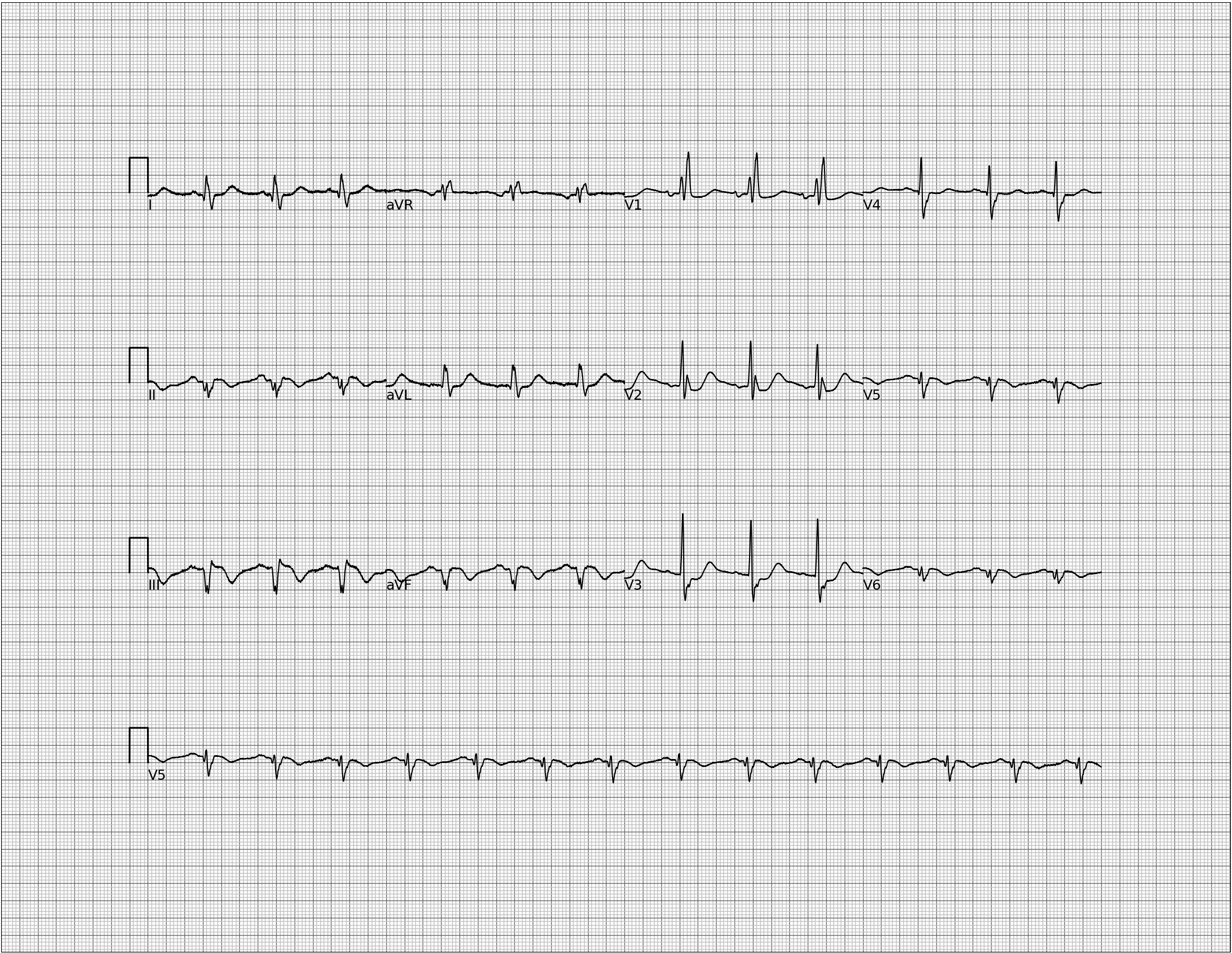}
         \caption{With lead names}
         \label{fig:ECG_with_lead_names}
     \end{subfigure}
     \hfill
     \begin{subfigure}[b]{0.49\textwidth}
         \centering
         \includegraphics[width=\textwidth]{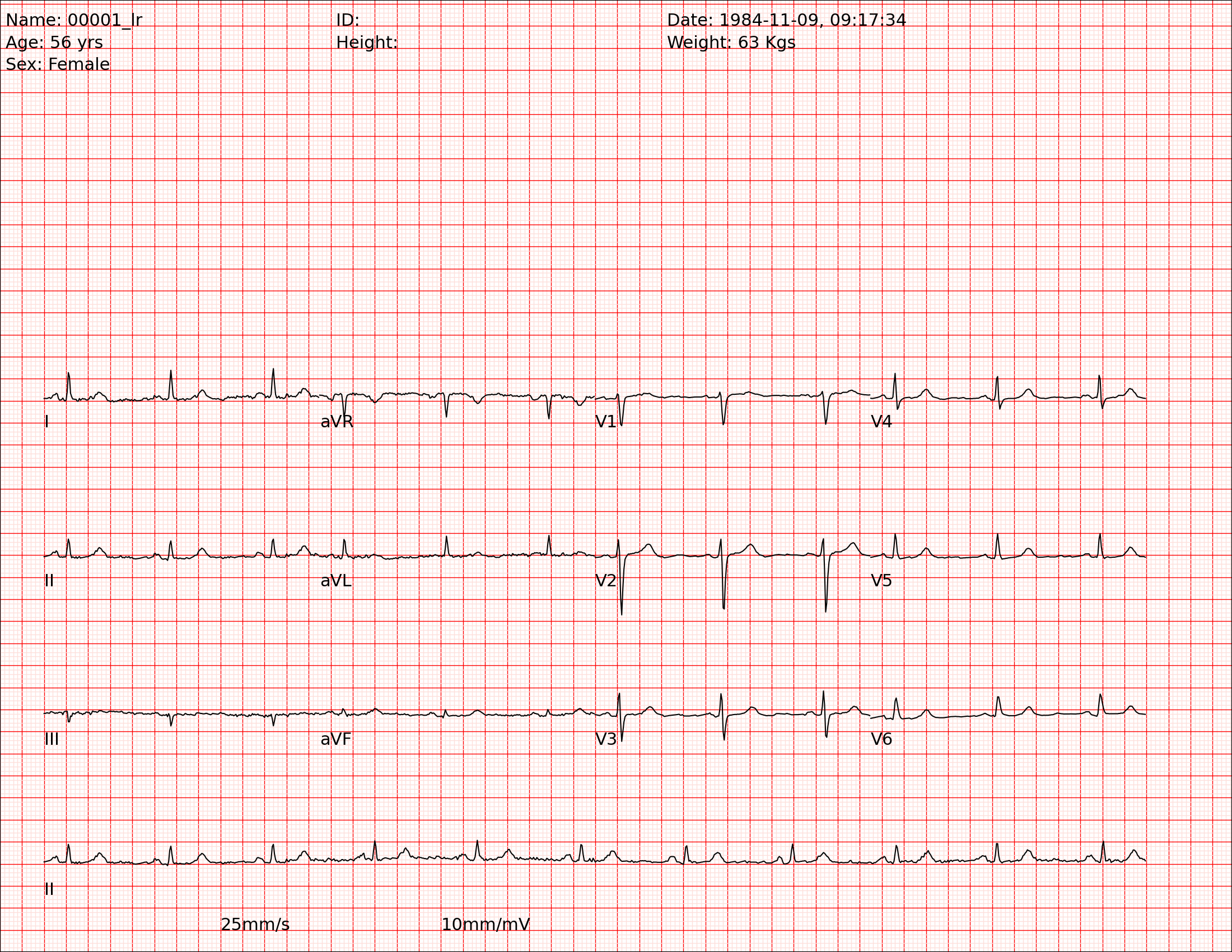}
         \caption{With text artifacts}
         \label{fig:ECG_with_text}
     \end{subfigure}
        \caption{Distortion-less synthetic ECG images with lead names (left) and printed text (right)}
        \label{fig:ECG_distortion_less}
\end{figure}

Typically, ECG records, whether in printed format or in EHRs, contain printed lead names, patient information/ID, ECG calibration pulse, date, physician/referrer's name, diagnostic codes/keywords, ECG-based measurements, and various medical terminologies. Many of these fields are protected health information (PHI) and need to be redacted to preserve privacy when shared in the public domain. ECG-Image-Kit accommodates the printing of such information through its command-line options. The lead names (I, II, III, V1, V2, V3, V4, V5, V6, aVL, aVF, aVR) are printed alongside their respective ECG segments on the synthetically generated ECG image (cf. Fig.~\ref{fig:ECG_with_lead_names}). The user of the toolbox has the option to choose whether there should be an overlap between ECG segments and printed text artifacts. Although overlapped characters pose a problem in digitizing paper ECG records \cite{ganesh2021combining}, they are added to represent realistic paper ECGs, which occasionally print text with partial overlap with the ECG traces. Further, to add other printed information such as date, patient record numbers, etc., the toolkit uses the corresponding fields from the WFDB header files that accompany all PhysioNet data files, or through a customizable text-based template file. These texts are superimposed on the synthetically generated ECG image obtained from the previous step (cf. Fig.~\ref{fig:ECG_with_text}).

% Insert image with 

% A distortionless ECG generated by ECG-Image-Kit ...

\subsubsection{Handwritten text artifacts}
Scanned ECGs may contain annotations or handwritten diagnoses from the healthcare provider. Our synthetic ECG image generation pipeline optionally simulates such handwritten text artifacts to create more realistic ECG images.

Most handwritten text on paper ECG records consists of medical terms related to cardiology. We collected a set of medical texts related to ECG and CVDs and used natural language processing (NLP) to extract relevant keywords and phrases. The resulting set of keywords and phrases was converted to handwritten-style images using pretrained models, and the resulting images were overlaid on the ECG images from the previous step of our pipeline. We used the Python-based \texttt{en\_core\_sci\_md} model from sciSpacy \cite{neumann2019scispacy} for the NLP step, which provides a fast and efficient pipeline for tokenization, parts-of-speech tagging, dependency parsing, and named entity recognition. Next, we retrained the SpaCy model \cite{honnibal2017spacy} on our collected medical texts to retain words/phrases in the ECG context. The dependency parser and the parts of speech tagger in the released models were retrained on the treebank of McClosky and Charniak \cite{mcclosky2008self}, which is based on the GENIA~1.0 corpus \cite{kim2003genia}. Major named entity recognition models was trained on the MedMentions dataset \cite{mohan2019medmentions}. ECG-Image-Kit parses words from an input text file or from online links using the BeautifulSoup library \cite{richardson2007beautiful}, performs parts-of-speech tagging on the parsed words, and then uses named entity recognition from the aforementioned models to identify ECG-related keywords, which are randomly chosen and added as handwritten text.

\begin{figure}[tb]
     \centering
     \begin{subfigure}[b]{0.49\textwidth}
         \centering
         \includegraphics[trim={0 3.8cm 0 0},clip,width=\textwidth]{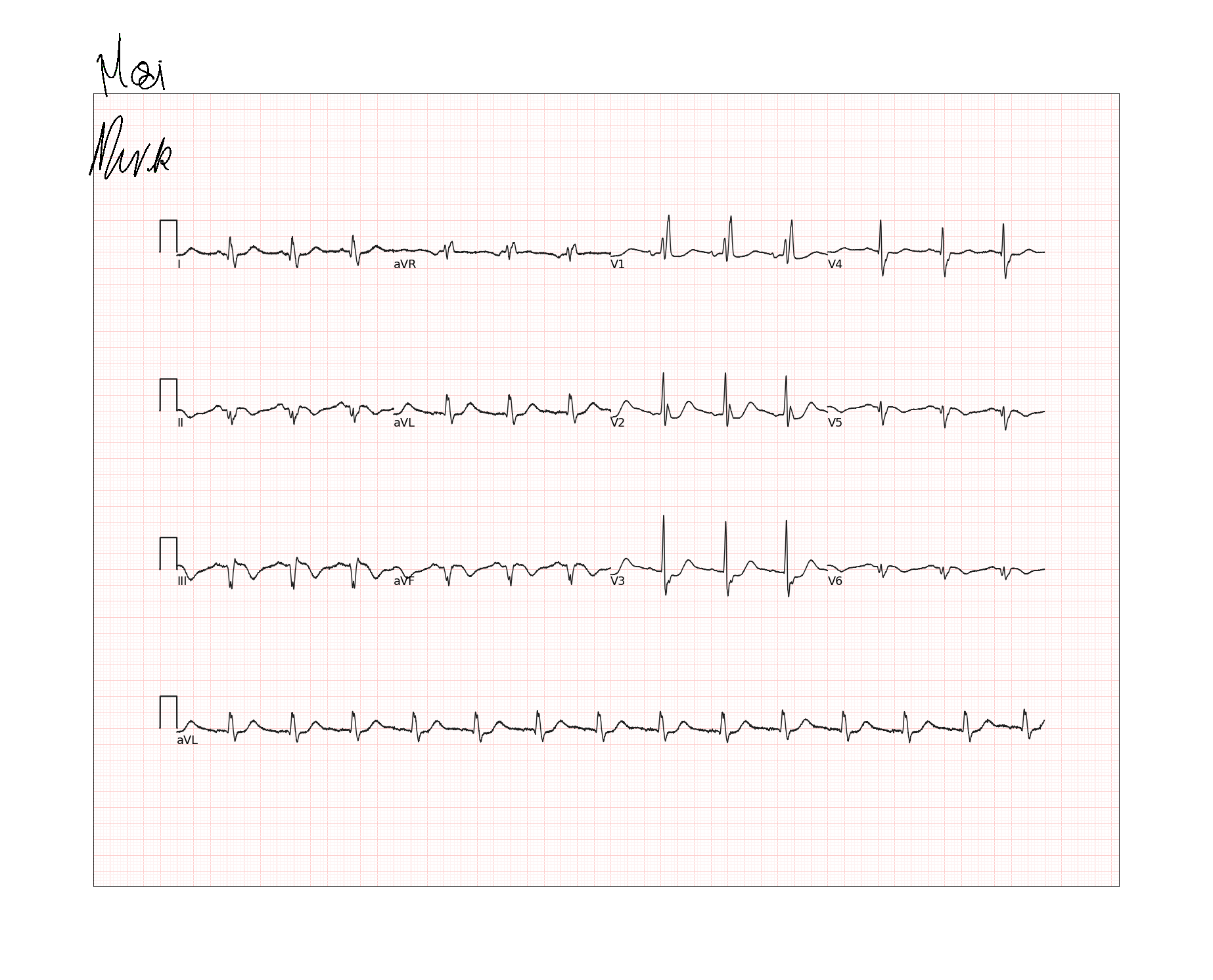}
         \caption{Handwritten-like artifacts}
         \label{fig:Handwritten1}
     \end{subfigure}
     \hfill
     \begin{subfigure}[b]{0.49\textwidth}
         \centering
         \includegraphics[trim={0 5cm 0 0},clip,width=\textwidth]{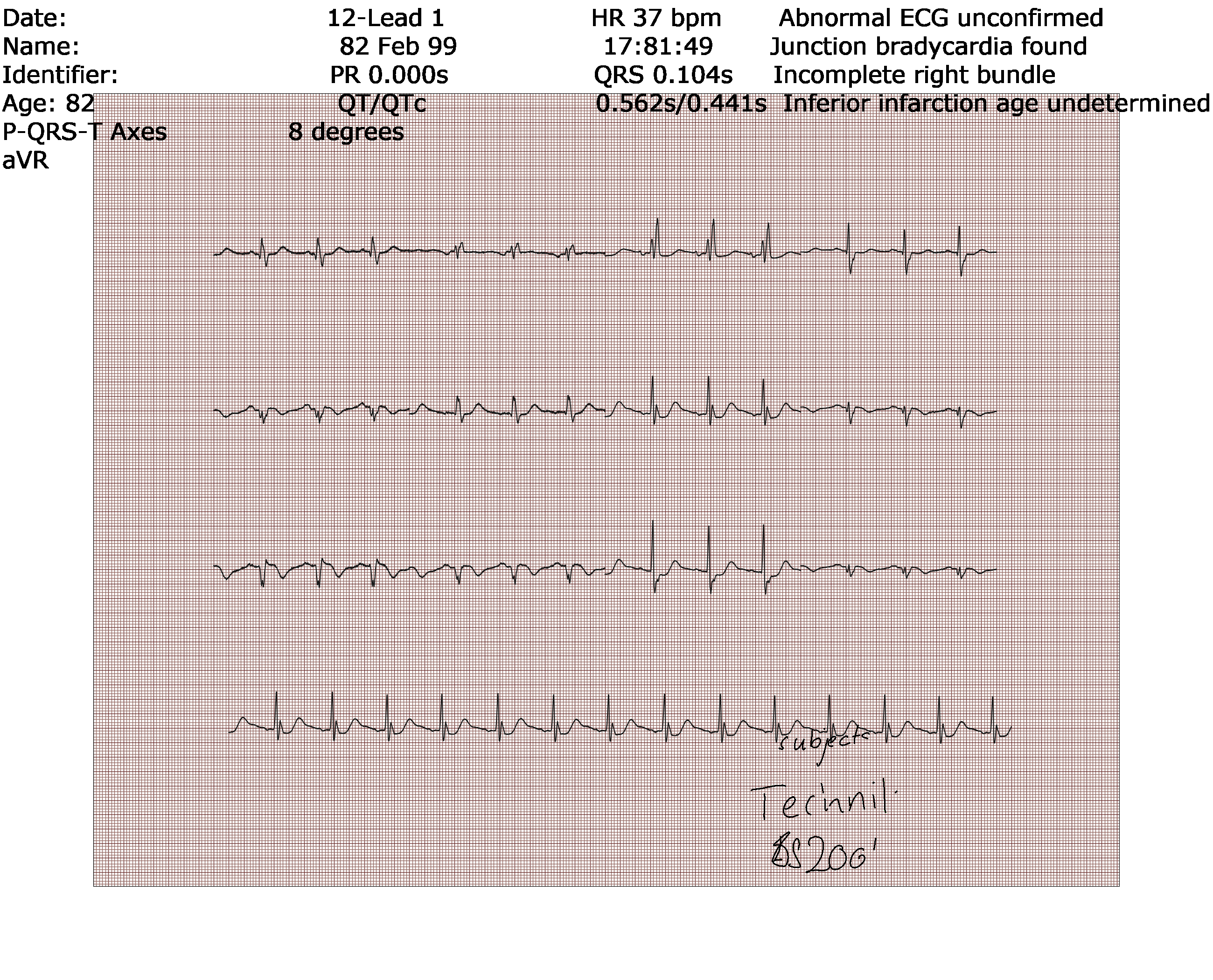}
         \caption{Printed text artifacts}
         \label{fig:Handwritten2}
     \end{subfigure}
        \caption{Handwritten and printed text artifacts on synthetic paper ECG}
        \label{fig:ECG_handwritten_dist}
\end{figure}

The extracted words are converted into handwritten-style text to overlay on the synthetic ECG image. We use a pretrained recurrent neural network (RNN) transducer-based model paired with a soft window to generate handwritten text from the extracted words \cite{graves2013generating}. One of the major challenges in converting words to handwritten text is that the input and output sequences vary greatly in length depending on the handwriting style, pen size, etc. The RNN transducer-based model can predict output sequences of different lengths and unknown alignments from the input sequence \cite{graves2012sequence}. The soft window determines the length of the output handwritten sequence by convolving with the input text string, resulting in outputs of varying lengths for different handwriting styles. Our current handwritten text pipeline allows the user to choose from seven different handwriting styles to overlay onto the ECG image. The coordinates for overlaying the text can be chosen by the user of the toolbox or, if not specified, are selected randomly. Examples of the resulting images are shown in Fig.~\ref{fig:ECG_handwritten_dist}.

\subsubsection{Paper wrinkles and creases}
Wrinkles and creases are common distortions in scanned paper-based ECG images. Wrinkle distortions arise due to the uneven surface of the wrinkled document. When the scanner light passes over this uneven surface, shadows or reflections from the wrinkles may distort the resulting image. This can cause areas of the image to appear darker or lighter than the actual paper, or lead to the image appearing blurry or distorted. Creases, in contrast, result from the physical fold in the scanned paper. As the scanner light passes over the crease, it can create a shadow that manifests as a dark/bright line in the resulting image, potentially making the text or image in the creased area difficult to read or interpret.

Creases can be simulated in images through image processing techniques, using blurred lines spaced linearly to represent paper folds. The placement and orientation of these crease artifacts are mathematically determined by line equations and translations, based on the crease count, inclination angle, and image dimensions. We apply Gaussian blurring to the crease lines to simulate blurring effects, commonly used in image processing to replicate the impact of an unfocused camera, that affect deep neural network performance \cite{dodge2016understanding}. This technique realistically integrates creases within image boundaries, enhancing synthetic paper-like ECG image authenticity. The blurring is mathematically represented as a convolution with a Gaussian kernel:
\begin{equation}
G(x, y) = \displaystyle\frac{1}{{2\pi\sigma^2}} e^{-\frac{x^2 + y^2}{2\sigma^2}}
\end{equation}
where $x$ and $y$ represent the coordinates in the Euclidean space, and $\sigma$ denotes the standard deviation of the Gaussian distribution in pixels. Applying Gaussian blurring to the crease lines results in more realistic images by simulating a shadow effect in the creases, commonly seen in scanned paper ECG images. Assuming a Gaussian mask of size $L\times L$, if the Gaussian kernel is centered within the mask, the blurring effect resembles light reflections off the top of the paper. Displacing the Gaussian kernel's center from the mask's center, resulting in non-symmetric blurring at each point, simulates light reflection from different angles. 

Wrinkles can be considered textures and synthesized using advanced texture synthesis techniques such as \textit{image quilting} \cite{efros2001image}. Image quilting begins with a plain wrinkle image as a seed, followed by the random selection of a patch from this image. This patch forms the foundation for synthesizing the entire wrinkle texture. Multiple patches are generated and seamlessly blended using the minimum boundary error cut method \cite{liang2001real}. This method aims to identify the optimal boundary between two patches by minimizing the error in the overlap region. The minimum cost path through the overlap is determined using dynamic programming based on the algorithm proposed in \cite{davis1998mosaics}. Rather than using a straight line between the patches, this method computes the minimum cost path that delineates the boundary of the new block. Consequently, the placement of texture patches appears smoother and more natural, significantly reducing noticeable sharp edges between the patches.

Let $B_{\text{1}}$ and $B_{\text{2}}$ be two blocks overlapping along their vertical edges, with $B_{\text{ov1}}$ and $B_{\text{ov2}}$ representing the regions of overlap between them. The error surface $e$ is defined as the squared difference between $B_{\text{ov1}}$ and $B_{\text{ov2}}$: $e = (B_{\text{ov1}} - B_{\text{ov2}})^2$. The error surface $e$ is traversed for each row $(i = 2,\ldots, N)$ in the overlapping region, and the minimum error path $E$ is computed using dynamic programming as follows:
\begin{equation}
E_{i, j} = e_{i, j} + \min(E_{i-1, j-1}, E_{i-1, j}, E_{i-1, j+1})
\end{equation}
where $E_{i, j}$ denotes the cumulative minimum error at position $(i, j)$ in the error matrix $E$. The minimum error path is obtained by taking the minimum of the three adjacent pixels in the previous row $(i-1)$ and adding it to the corresponding pixel in the current row $(i)$. The last row of $E$ contains the minimal value, indicating the end of the minimal vertical path through the error surface. Backtracking from the last row identifies the path of the best-fit boundary between $B_{\text{1}}$ and $B_{\text{2}}$ \cite{efros2001image}, resulting in a seamless and visually coherent texture synthesis. This image quilting technique, combined with the minimum boundary error cut method, helps generate a realistic paper-based texture with wrinkles and creases. Thus the resultant image exhibits natural and realistic blending, contributing to the overall authenticity of the synthetic ECG images.

\begin{figure}[tb]
     \centering
     \begin{subfigure}[b]{0.45\textwidth}
         \centering
         \includegraphics[width=\textwidth]{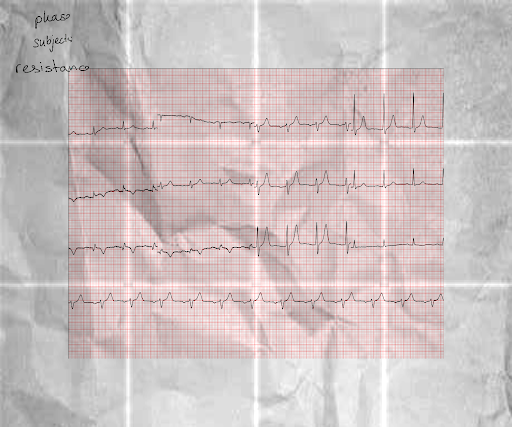}
         \caption{Wrinkles and creases}
         \label{fig:wrinkle1}
     \end{subfigure}
     \hfill
     \begin{subfigure}[b]{0.45\textwidth}
         \centering
         \includegraphics[width=\textwidth]{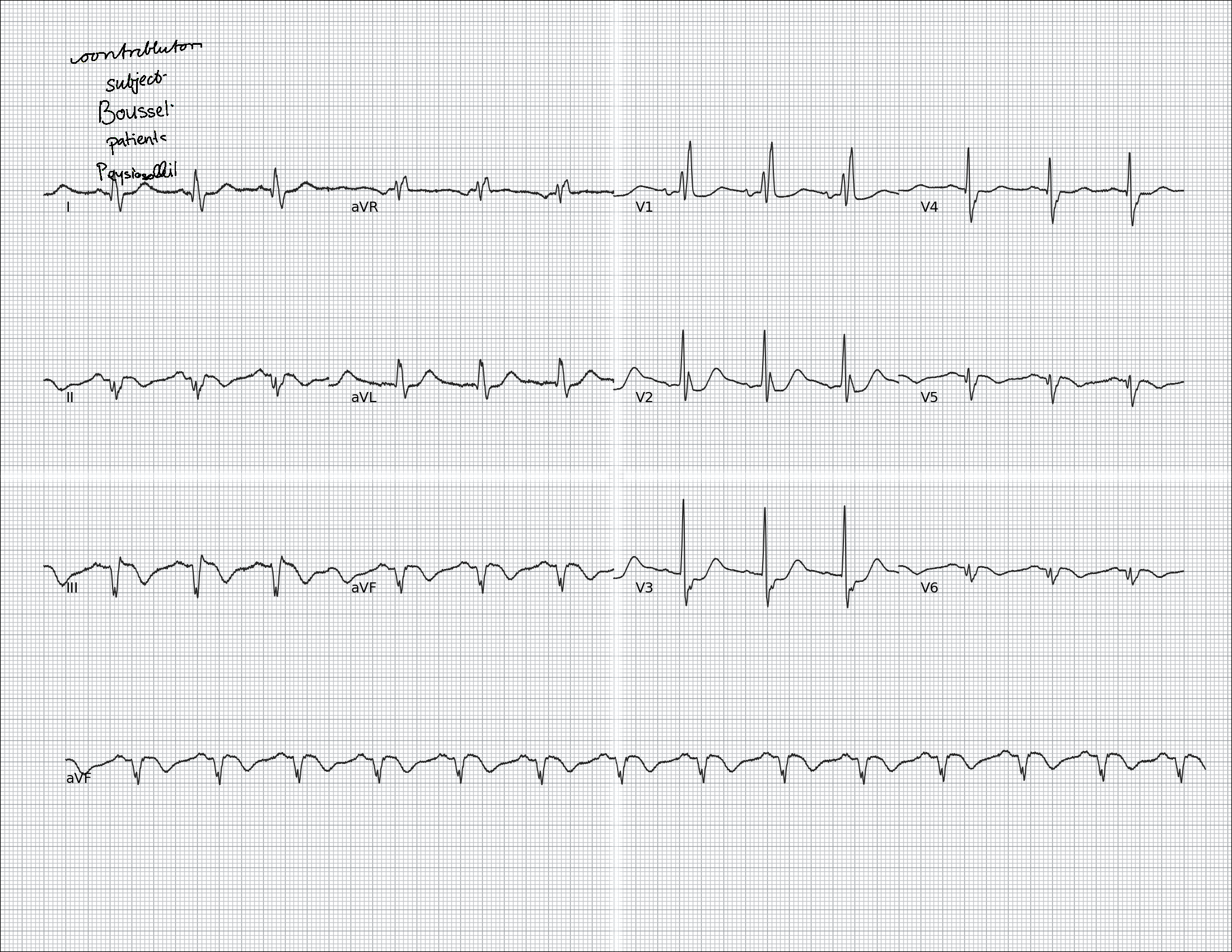}
         \caption{Crease artifacts}
         \label{fig:wrinkle2}
     \end{subfigure}
        \caption{Wrinkle and crease artifacts on synthetic ECG images}
        \label{fig:ECG-wrinkle-examples}
\end{figure}

Together, wrinkles and creases are added as cumulative transforms on the ECG image to add realistic distortions, as shown in Fig.~\ref{fig:ECG-wrinkle-examples}.
\subsubsection{Perspective artifacts}
We apply perspective transformations on the synthetic paper ECG image to incorporate different camera viewpoints that could have been used while scanning or taking pictures of paper ECG records. Traditionally, perspective transformations have been widely used in image augmentation for computer vision applications. \cite{wang2019perspective} refers to the use of perspective transformations for data augmentation to produce new images captured from different camera viewpoints, specifically for object detection applications. Perspective transforms are simulated by applying geometric transformations to distortionless ECG images. Affine, projective, or homography transformations are utilized to introduce variations in scaling, rotation, and skewing, mimicking the perspective distortions encountered in real paper ECGs. Affine transformations preserve parallelism of lines and can be used to simulate translations, scaling, rotations and shear transformations. Thus, they simulate parallel movements of a camera when scanning a paper ECG. Given an ECG image, the affine transformation can be represented by the matrix transform:
\begin{equation}
\left( \begin{array}{c}
x' \\
y' \\
1 \\
\end{array} \right)
=
\left( \begin{array}{ccc}
a & b & c \\
d & e & f \\
0 & 0 & 1 \\
\end{array} \right)
\left( \begin{array}{c}
x \\
y \\
1 \\
\end{array} \right)
\end{equation}
where $(x, y)$ are the coordinates of a pixel in the original image $I$, and $(x', y')$ are the transformed coordinates after the affine transformation. 

Projective transformations, which include affine transformations as a special case, can also simulate skew transformations. Unlike affine transformations, projective transformations do not preserve parallelism and are employed to simulate the alteration of perceived images with changes in the observer's viewpoint. In ECG-Image-Kit, projective transformations are integrated into the synthetic paper ECG generation pipeline to mimic observational changes in mobile- or camera-based ECG images. The matrix equation for a projective, or perspective, transformation is given by:
\begin{equation}
\left( \begin{array}{c}
x' \\
y' \\
w' \\
\end{array} \right)
=
\left( \begin{array}{ccc}
a & b & c \\
d & e & f \\
g & h & i \\
\end{array} \right)
\left( \begin{array}{c}
x \\
y \\
1 \\
\end{array} \right)
\label{eq:projective_map}
\end{equation}
where $(x, y)$ are the coordinates of a pixel in the original ECG image $I$, $(x', y')$ are the transformed coordinates after the projective transformation, and $w'$ is a scaling factor that ensures homogeneous coordinate representation. In \eqref{eq:projective_map}, the elements $a$, $b$, $c$, $d$, $e$, $f$, $g$, $h$, and $i$ control the scaling, rotation, shear, skew and perspective effects applied to the image. These transformations add depth and simulate different viewing angles and positions, further enhancing the resemblance to real ECG images. 

The aforementioned transformations are implemented using the \textit{imgaug} library in Python \cite{jung2019imgaug}, a tool for image augmentation in machine learning experiments. Imgaug supports numerous augmentation techniques, enables their easy combination, and executes them in random order, making it ideal for generating highly variable synthetic ECG image datasets.

\subsubsection{Imaging artifacts and noise}
The final distortions added to the synthetic paper ECG images include generic imaging noise modeled by Gaussian noise, Poisson noise, salt and pepper noise, and color temperatures. These are crucial for creating realistic synthetic images and enhancing the robustness of machine and deep learning models trained on these images. Gaussian noise typically arises in digital images during image acquisition, which in this case involves scanning or photographing paper ECG images or ECG images from a monitor. Modeling sensor noise, Gaussian noise is added independently to each pixel:
\begin{equation}
I_{\text{noisy}}(x, y) = I(x, y) + n_{\text{gaussian}}(0, \eta)
\end{equation}
where $I_{\text{noisy}}(x, y)$ represents the noisy pixel value, $I(x, y)$ is the original pixel value of the ECG images, and $n_{\text{gaussian}}(0, \eta)$ is a random value drawn from a Gaussian distribution with mean 0 and standard deviation $\eta$.

Poisson-distributed Shot noise, commonly used to model electromagnetic and photonics noises during image acquisition, can be added to each pixel:
\begin{equation}
I_{\text{noisy}}(x, y) = \min\{255, \max[0, I(x, y) + n_{\text{poisson}}(\lambda)]\}
\end{equation}
where $n_{\text{poisson}}(\lambda)$ is a random value drawn from a Poisson distribution with parameter $\lambda$, and the clipping ensures the pixel value remains within the range [0, 255] (for 8-bit per color image representations).

Salt and pepper noise models camera sensor malfunctions, which may occur when scanning ECG images. This noise randomly sets pixels to either the minimum or maximum intensity:
\begin{equation}
I_{\text{noisy}}(x, y) = \begin{cases}
0, & \text{with probability } p/2 \\
255, & \text{with probability } p/2 \\
I(x, y), & \text{with probability } 1 - p
\end{cases}
\end{equation}
where the probability $p$ determines the density of the salt and pepper noise.

Finally, color temperatures are simulated by adjusting the image's color channels. A color temperature value is selected, and the color channels are transformed using algorithms such as color temperature scaling or white balance adjustment to mimic the desired effect. The RGB values of the image change according to a temperature value ranging from 1000 to 40000 Kelvin, with lower values corresponding to bluish tinges and higher values to orangish tinges. Modifications to color temperature are useful for simulating the aging and wearing effects of ECG thermal paper.

In ECG-Image-Kit, these artifacts are added in user-adjustable proportions to the synthetic ECG image using the imgaug library in Python.

\section{Case Study: A combined image processing and deep learning model for ECG image digitization}
We utilized ECG-Image-Kit to create a diverse dataset for training an ECG digitization model. These synthetic ECG images were generated from the PTB diagnostic ECG database time-series consisting of 549 records \cite{bousseljot1995nutzung}. The trained model was then applied to evaluate the fidelity of a synthetic paper ECG dataset generated from the PhysioNet QT Database \cite{laguna1997database,PhysioNetQT,Goldberger2000}, both in terms of signal quality and accuracy of extracting ECG-based measurements.
%%%% PTB-XL dataset \cite{wagner2020ptb}, which includes 1000 images. It was also used to assess the accuracy of ECG-based RR- and QT-interval measurements of the PhysioNet QT Database \cite{laguna1997database,PhysioNetQT}.

The digitization process involves multiple steps. First, preprocessing techniques including image registration and optical character recognition (OCR) are used to correct distortions and remove text from the images. Next, a denoising CNN network, trained on the synthetic dataset, denoises the images and eliminates the ECG grid. The denoised image is then divided into segments, processed to address discontinuities, and transformed into corresponding time-series data for ECG-based waveform measurements. Each step of the ECG digitization pipeline is detailed in this section.

\subsection{Rotation compensation}
\label{sec:rotation_compensation}
Rotation compensation is a crucial preprocessing step for images captured by cameras or scanned with minor or major rotations. This step aims to align the images and remove any skew, shear, and rotations in the scanned paper ECG images. We explored two methods for rotation compensation: keypoint matching and a Radon transform-based technique. In our experiments, the Radon transform-based method proved to be more effective for rotation compensation.

\subsubsection{Image registration using keypoint matching}
Image registration involves matching, aligning, and overlaying two or more images of a scene captured from different viewpoints. It transforms different image sets into a single unified coordinate system and is widely used in vision-based applications \cite{8346440}. The main steps in image registration include keypoint detection, keypoint matching, and image reconstruction from keypoints. Keypoints are specific points that characterize the image and are used to compute the transformation. Descriptors, which are histograms of the image gradient, characterize the appearance of a keypoint. The Oriented FAST and Rotated BRIEF descriptor (ORB) algorithm and the Scale-Invariant Feature Transform (SIFT) algorithm are two methods for keypoint detection. ORB is notably more efficient and faster than SIFT, being computationally faster by nearly two orders of magnitude \cite{rublee2011orb}.

The ORB algorithm \cite{rublee2011orb} can be used to analyze a reference paper ECG image $I_{\text{ref}}$ and the input paper ECG image $I_{\text{input}}$ to identify regions of significant pixel intensity change. ORB uses a combination of oriented FAST (Features from Accelerated Segment Test) for corner detection and BRIEF (Binary Robust Independent Elementary Feature) for descriptor computation. The oFAST algorithm detects keypoints in both $I_{\text{ref}}$ and $I_{\text{input}}$, ordering them using the Harris Corner measure \cite{rosin1999measuring}. To ensure scale invariance of detected keypoints, ORB employs a multi-scale image pyramid, with each level comprising a downsampled version of the image from the previous level. By detecting keypoints at each level, ORB effectively locates important points across different scales. After obtaining the keypoints, the orientation for each keypoint patch is computed. rBRIEF (Rotation-aware BRIEF) converts the keypoints identified by oFAST into a binary feature vector. The descriptors, $d_{\text{ref}}'$ and $d_{\text{input}}'$, are calculated and compared to establish correspondences between keypoints in the reference and input images, thereby generating the rotation matrix to align the input image with the reference axes.

However, not all detected keypoints may have reliable matches due to noise or image artifacts. To mitigate this, we use the Random Sample Consensus (RANSAC) algorithm, a robust matching technique \cite{VINAYA2015174}. RANSAC iteratively selects a subset of correspondences from the set $C$, denoting correspondences between keypoints in $I_{\text{ref}}$ and $I_{\text{input}}$ as $(k_{\text{ref},i}, k_{\text{input},i})$, and estimates the transformation matrix $T$ that best aligns the keypoints $k_{\text{ref}}'$ and $k_{\text{input}}'$. This process, repeated multiple times, refines the transformation estimate and filters out outliers. Once the affine transformation matrix $T$ is obtained, it is applied to the input scanned ECG image $I_{\text{input}}$, warping the image to match the spatial alignment of the reference image $I_{\text{ref}}$. This registration process ensures the input ECG image closely resembles the reference image in geometric configuration.

\begin{figure}[tb]
     \centering
     \begin{subfigure}[b]{0.48\textwidth}
         \centering
         \includegraphics[width=\textwidth]{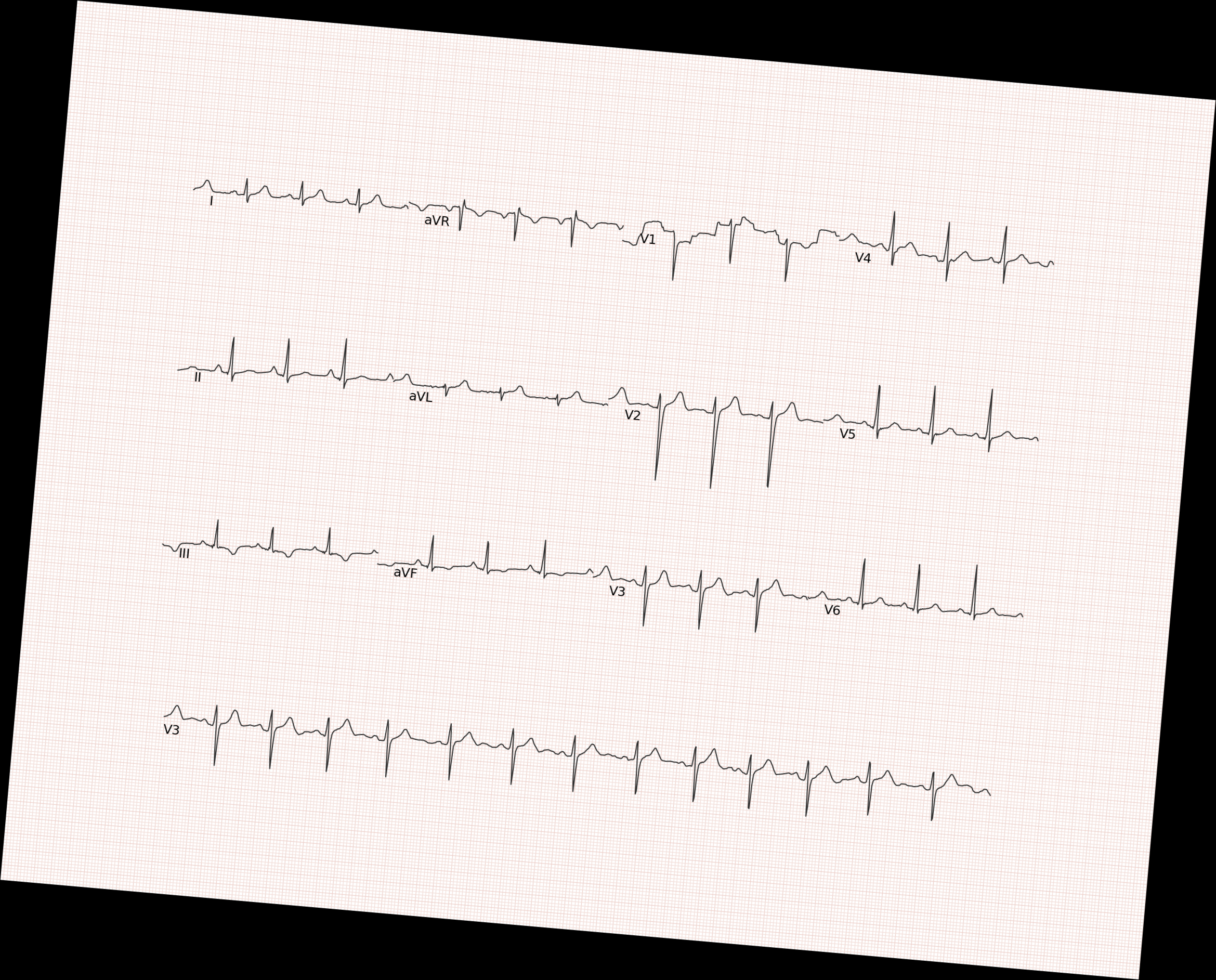}
         \caption{With rotations}
         \label{fig:ORB_rotated}
     \end{subfigure}
     \hfill
     \begin{subfigure}[b]{0.48\textwidth}
         \centering
         \includegraphics[width=\textwidth]{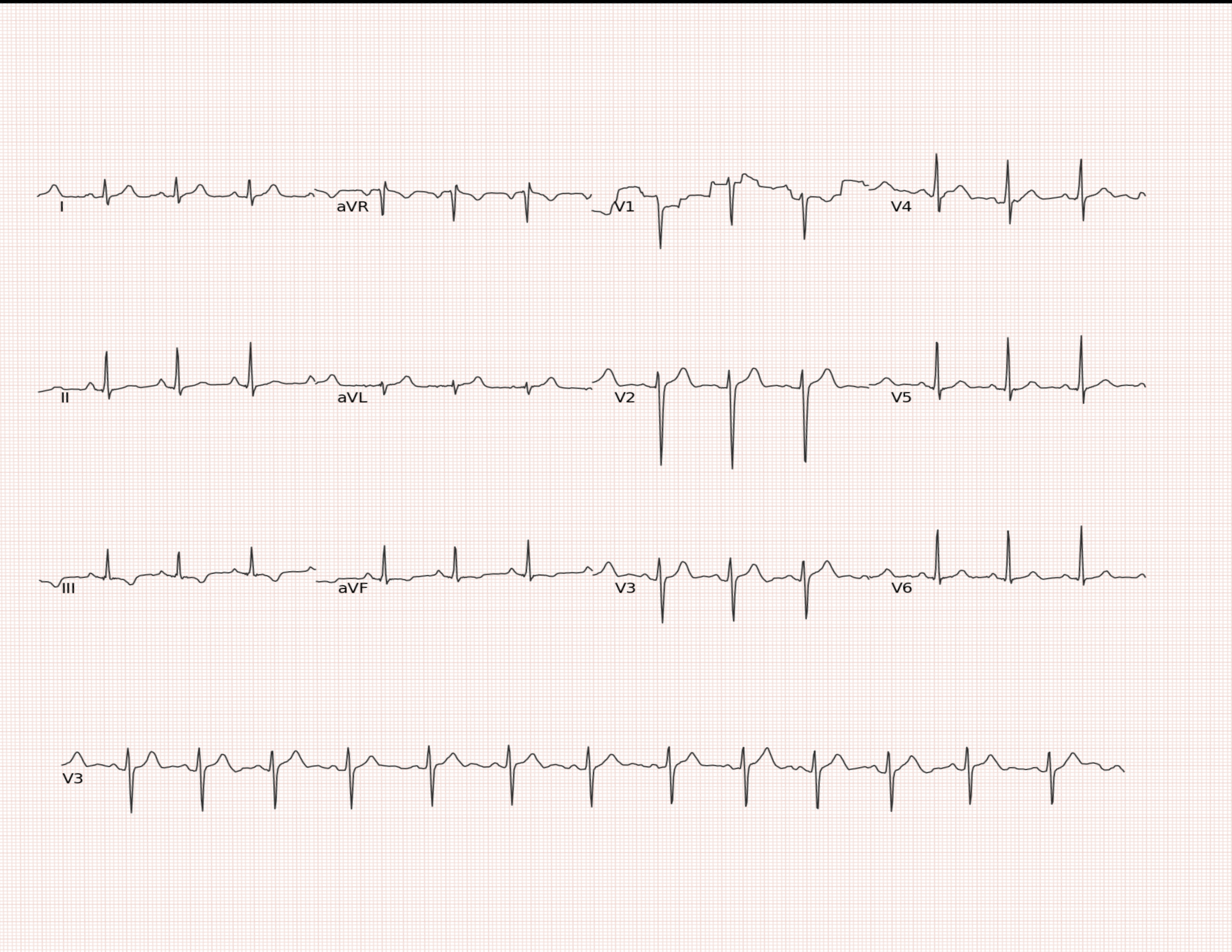}
         \caption{Post rotation compensation}
         \label{fig:post-ORB}
     \end{subfigure}
        \caption{Image alignment using Oriented FAST and Rotated BRIEF descriptor (ORB)}
        \label{fig:ORB}
\end{figure}

\begin{figure}[tb]
     \centering
         \includegraphics[width=\textwidth]{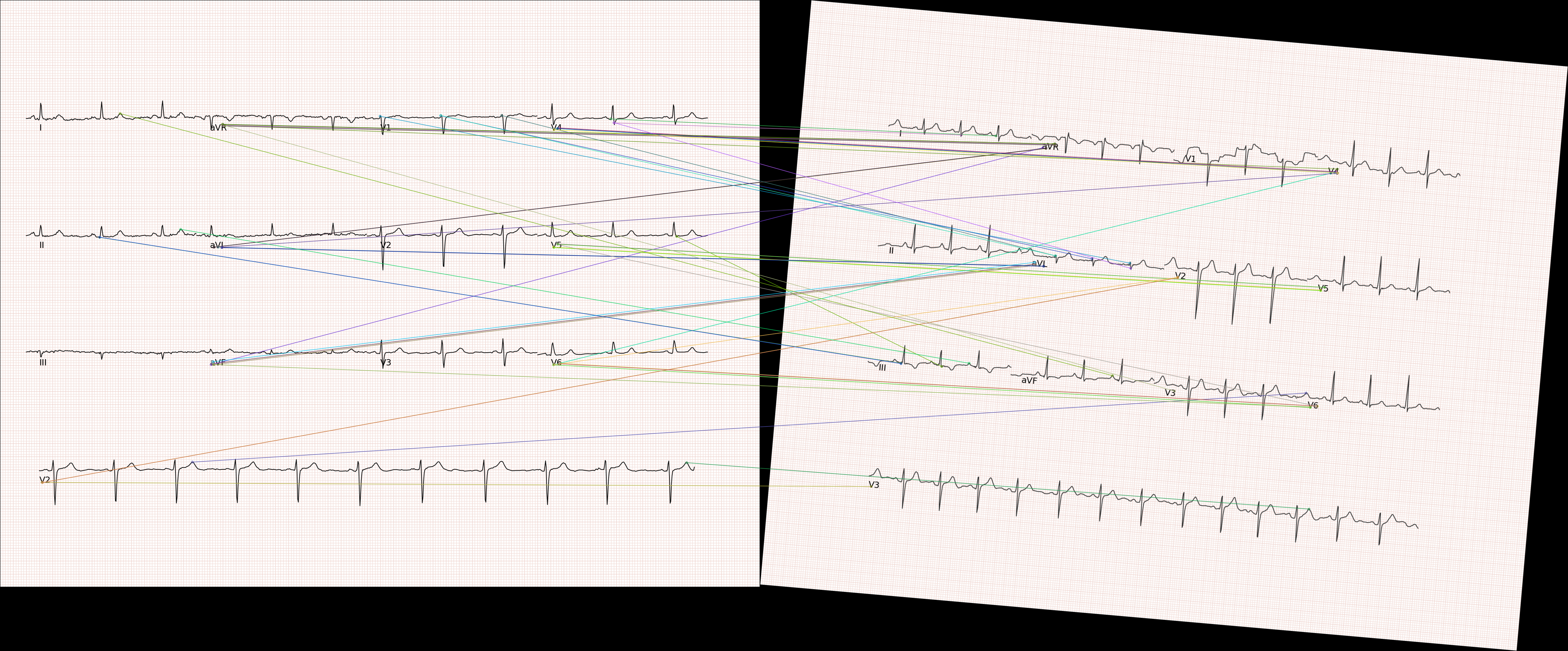}
         \caption{Matched keypoints in rotated image and reference image}
         \label{fig:matching_keypoints}
\end{figure}

In the use case, we employed ORB-RANSAC to detect 50 keypoints in the ECG image and a template ECG image. After computing keypoints, we used a greedy algorithm for point matching between the two images, employing the Hamming distance as the metric for keypoint matching. We selected the match corresponding to the least Hamming distance. Subsequently, we determined the homography matrix based on the best-matched keypoints and applied this matrix to the rotated ECG image to restore the original image and compensate for rotation. However, the ORB algorithm may not work for all scanned paper ECG images due to high variability in the dataset, making it challenging to compute keypoints from a template ECG image. Sample results obtained from ORB are shown in Figs.~\ref{fig:ORB} and \ref{fig:matching_keypoints}.

\subsubsection{Image registration using Radon transform}
The Radon transform is widely used for computed tomographic reconstruction \cite{helgason1980radon}. It mathematically represents an image in terms of its projection profiles. In continuous form, the Radon transform of a 2D signal (or image) $I(x, y)$ is defined as:
\begin{equation}
R(\rho, \theta) = \int_{-\infty}^{\infty} \int_{-\infty}^{\infty} I(x, y) \cdot \delta(x\cos\theta + y\sin\theta - \rho) \, dx \, dy   
\end{equation}
where $\rho$ is the distance parameter, $\theta$ is the angle parameter, and $\delta(\cdot)$ represents the Dirac delta function.

The Radon transform's fundamental principle is that $(n-1)$-dimensional line integrals through an $n$-dimensional volume (like a 2D image) allow the recovery of original $n$-dimensional Fourier values through their $(n-1)$-dimensional Fourier transform. It transforms an $n$-dimensional volume into a complete set of $(n-1)$-dimensional line integrals. The image is sampled along a set of $(n-1)$ parallel lines at varying angles, accumulating intensity values along each line integral to form a projection profile. The Radon transform represents these profiles as a function of distance parameter $\rho$ and angle parameter $\theta$. The inverse Radon transform reverts these line integrals to the original image \cite{10.1145/3506651.3506654}. The Radon transform plot across different angles, known as a Sinogram due to its sinusoidal shape, reveals the image's rotation angle. Consequently, inversely rotating the image by the angle estimated from the Radon transform restores the original ECG image.

The Radon transform is particularly useful for tasks like rotation compensation \cite{bisht2014image}, \cite{nacereddine2015similarity}. By analyzing projection profiles at various angles, it enables the estimation of rotation angles between two images or signals. This facilitates the alignment or correction of rotation in images or signals, aiding various image processing and analysis tasks, such as correcting rotation or shear transformations in scanned images.

\subsection{Character removal}
Text artifact removal is the next step in our pipeline, ensuring the accurate digitization of underlying ECG signals. Character removal is achieved using standard OCR algorithms that execute text localization and detection. This process is followed by image inpainting to mask the text present in the image.

Text localization is the initial step in our approach, aiming to identify and localize text regions within the scanned ECG image. This algorithm detects regions containing characters but does not recognize individual characters. The Keras-OCR library utilizes CRAFT (Character Recognition Awareness For Text detection) for text localization. CRAFT employs a fully convolutional network architecture based on VGG-16 for encoding, and its decoding part includes skip connections similar to U-Net \cite{8953846}. Additionally, CRAFT has a refined anchor box generation scheme to predict text regions, providing bounding boxes around areas expected to contain characters. The algorithm generates a set of bounding boxes, denoted as $B = \{B_i\}$, where each bounding box $B_i$ is represented by its top-left and bottom-right coordinates $(x_1, y_1)$ and $(x_2, y_2)$, respectively.

Following text localization, the CRNN model is employed for text detection within localized regions. Keras-OCR implements a CRNN-based model for text recognition. CRNN, combining convolutional neural networks and recurrent neural networks, processes images containing sequential information like letters \cite{shi2016end}. The CRNN model learns the mapping function $F$ to transform localized text regions into corresponding text labels, denoted as $L = \{L_i\}$, where each $L_i$ represents the recognized text in the $i$th region enclosed by bounding box $B_i$.

After obtaining the character mask, it is blanked out by applying image inpainting using the \textit{fast marching method}, a prevalent technique for this purpose \cite{telea2004image}. Image inpainting in removed text regions of the ECG image involves several steps. The fast marching method employs a distance map, indicating the distances from known or inpainted pixels to missing regions. This map directs the inpainting process, prioritizing pixels close to the missing text regions. For each missing pixel, a first-order approximation is calculated using nearby pixels, their image values, and gradients, following the equation: 
\begin{equation}
Iq(p) = I(q) + \nabla I(q)(p - q)  
\end{equation}
where $I(q)$ denotes the image value at pixel $q$, and $\nabla I(q)$ the gradient at pixel $q$. The distance map aids in selecting the most appropriate nearby pixels for the approximation based on their proximity to the missing pixels. Leveraging the distance map and accounting for local image structures and gradients, the fast marching method effectively inpaints the removed text regions in the ECG image.

\begin{figure}[tb]
     \centering
     \begin{subfigure}[b]{0.8\textwidth}
         \centering
         \includegraphics[trim={0cm 0 0cm 0},clip,width=\textwidth]{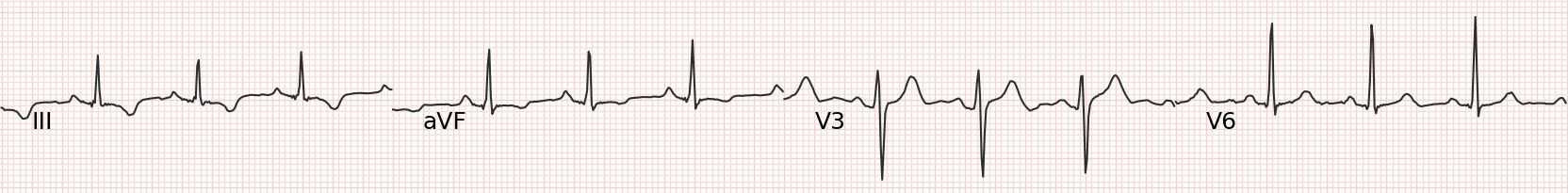}
         \caption{With text artifacts}
         \label{fig:text}
     \end{subfigure}
     \hfill
     \begin{subfigure}[b]{0.8\textwidth}
         \centering
         \includegraphics[trim={0cm 0 0cm 0},clip,width=\textwidth]{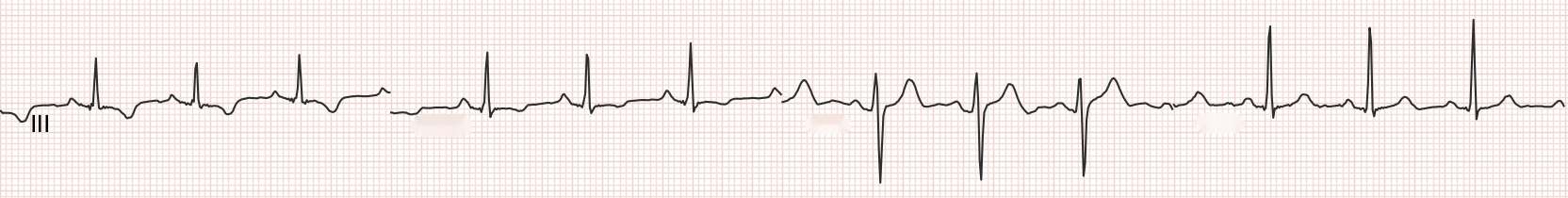}
         \caption{Post character removal}
         \label{fig:post-char-removal}
     \end{subfigure}
        \caption{Removal of text artifacts through optical character recognition}
        \label{fig:OCR}
\end{figure}

Overall, the combination of text detection, the fast marching method, and the use of a distance map facilitates accurate and efficient inpainting of removed text regions in the ECG image. An example of our text removal stage is illustrated in Fig.~\ref{fig:OCR}.

\subsection{Grid removal}
\begin{figure}[tb]
     \centering
         \includegraphics[width=.8\textwidth]{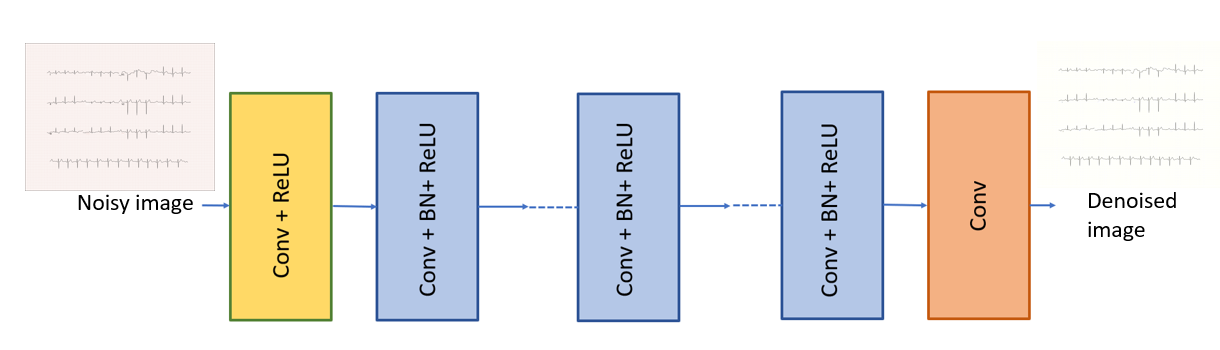}
         \caption{Denoising CNN architecture used for grid removal}
         \label{fig:DnCNN}
\end{figure}

\begin{figure}[tb]
     \centering
     \begin{subfigure}[b]{0.48\textwidth}
         \centering
         \includegraphics[width=\textwidth]{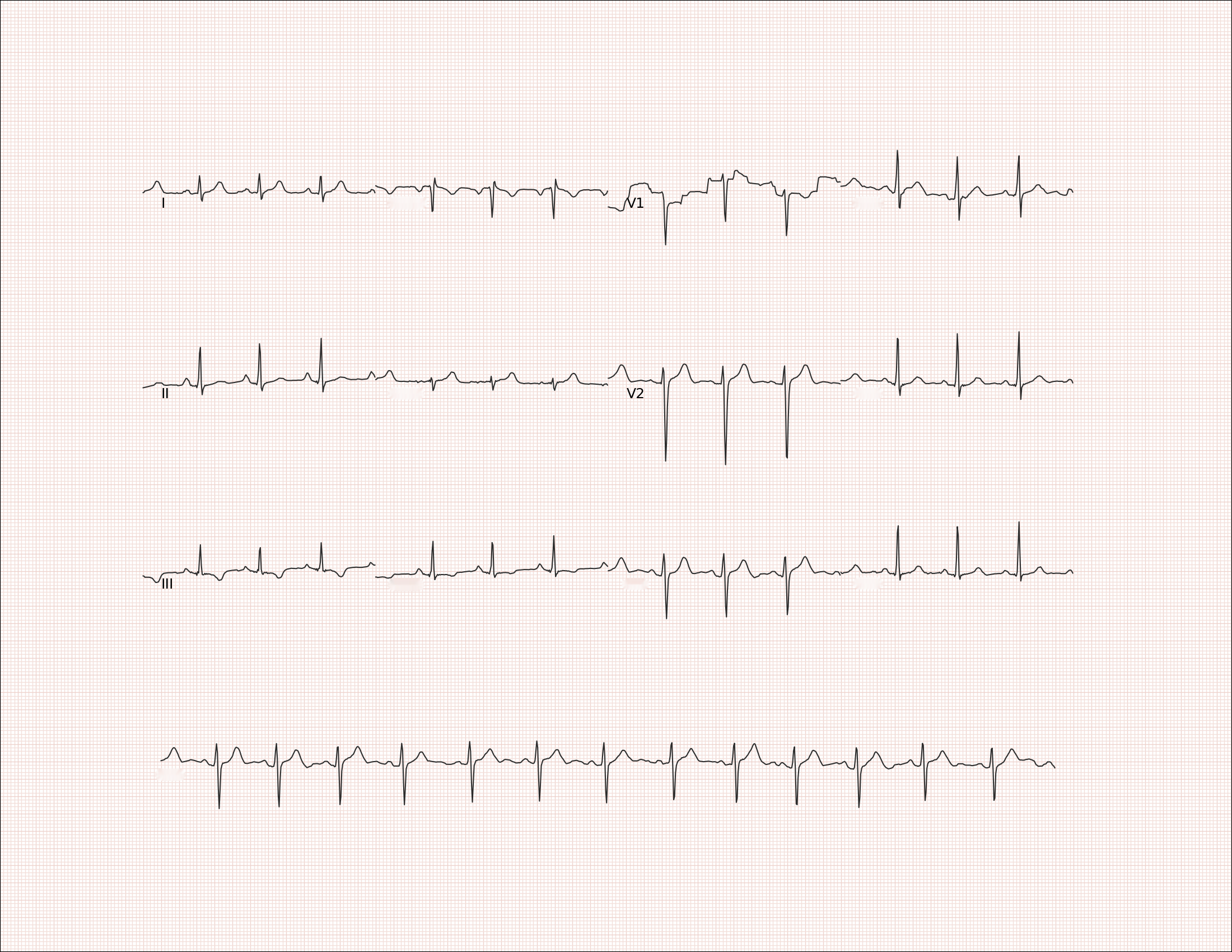}
         \caption{Before grid removal}
         \label{fig:before-gr}
     \end{subfigure}
     \hfill
     \begin{subfigure}[b]{0.48\textwidth}
         \centering
         \includegraphics[width=\textwidth]{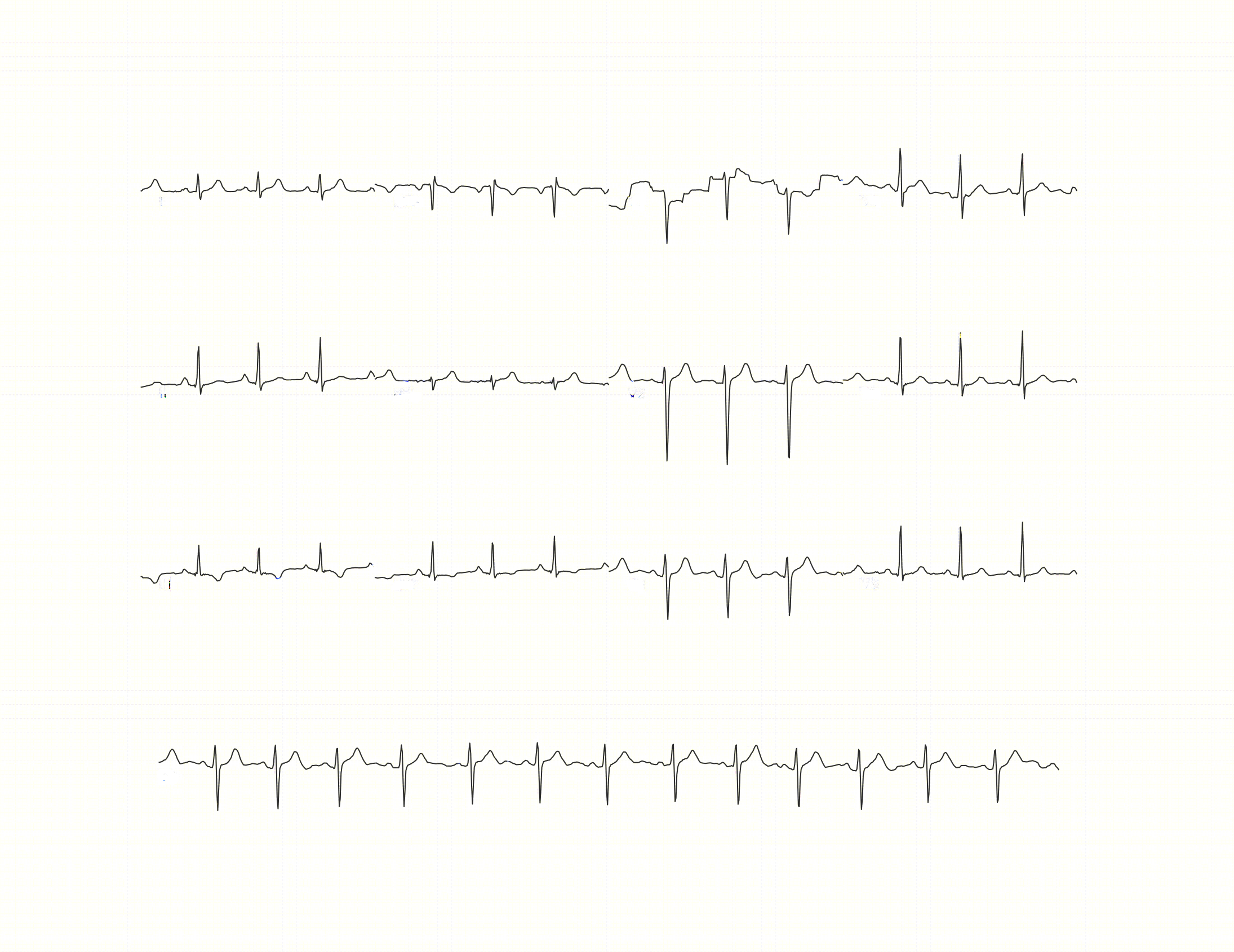}
         \caption{Post grid removal}
         \label{fig:After-gr}
     \end{subfigure}
        \caption{Synthetic paper ECG image before and after grid removal}
        \label{fig:grid}
\end{figure}

We formulated ECG paper grid removal as a denoising problem using the Denoising CNN (DnCNN) architecture (Fig.~\ref{fig:DnCNN}), which effectively handles Gaussian noise at unknown levels and manages three general image denoising tasks: blind Gaussian denoising, single-image super-resolution with multiple upscaling factors, and JPEG image deblocking with varying quality factors \cite{zhang2017beyond}. The input to DnCNN is a noisy observation $y = x + v$, where $x$ represents the clean ECG and $v$ denotes the background grid, and the expected output is a clean ECG plotted on a white paper background. The sample result is shown in Fig.~\ref{fig:grid}.

\subsubsection{Training the denoising CNN model}
Synthetic ECG images were generated using ECG-Image-Kit at 200$\pm$5 DPI resolution. The RGB images were divided into 3 channels, as the model will be trained on single-channel images. Each channel was further divided into 30$\times$30 pixel patches with a 5-pixel overlap. The raw image patches were scaled from 0-255 to 0-1. These patches, generated from a diverse dataset synthesized from 549 time-series records of the PTB Dataset \cite{bousseljot1995nutzung}, result in approximately 100,000 patches with corresponding ground truth for training. The patch dataset was next shuffled, and a 5-fold cross-validation was conducted for model evaluation. In each round, images in the leave-out set (20\%) were used solely for testing, and patches were generated from the 80\% training set, ensuring no testing image is leaked into the training set. This resulted in a 20:80\% test:train split. The patch dataset was then fed into the denoising neural network.

For the network architecture, we used a convolutional kernel of size 7$\times$7 to capture a significant portion of the grid in the patches \cite{simonyan2014very}. A high noise level typically requires a larger effective patch size to capture more context information for restoration \cite{levin2011natural}. Assuming the $y = x + v$ model, where $x$ represents the clean ECG and $v$ denotes the background grid, we can adopt a residual learning-based formulation to learn a residual mapping $\mathcal{R}(\mathbf{y}) \approx \mathbf{v}$. Here, the residual image to be predicted is considered to be the grid noise. The loss in terms of the trainable parameters is given by:
\begin{equation}
\ell(\boldsymbol{\Theta})=\frac{1}{2 N} \sum_{i=1}^{N}\left\|\mathcal{R}\left(\mathbf{y}_{i} ; \boldsymbol{\Theta}\right)-\left(\mathbf{y}_{i}-\mathbf{x}_{i}\right)\right\|_{F}^{2}    
\end{equation}
where $\boldsymbol{\Theta}$ represents the trainable parameters, $\left\{\left(\mathbf{y}_{i}, \mathbf{x}_{i}\right)\right\}_{i=1}^{N}$ represents $N$ noisy-ground truth patch pairs, and $\left\|\cdot\right\|_{F}$ is the Frobenius norm. The training goal is to learn $\mathcal{R}(\mathbf{y})$ and subtract this predicted residual from the image to obtain the clean ECG image without the grid.

The DnCNN we used comprises 17 layers. The first layer consists of a Conv2D + ReLU layer with 64 filters of size 7$\times$7. Rectified Linear Units were used for non-linearity in the architecture. By combining convolution with ReLU, DnCNN progressively separates image structure from noisy observation through hidden layers. Layers 2 to 16 consist of Conv2D + Batch normalization + ReLU layers with 64 filters of size 7$\times$7$\times$64. Batch normalization was done to accelerate training and enhance denoising performance \cite{zhang2017beyond}. The last layer used 1 filter of size 7$\times$7$\times$64 to produce the output image. The model was trained using early stopping for 30 epochs until mean squared error loss saturation, reducing the loss to an order of 10\textsuperscript{-5}. The model was also trained without early stopping for 500 epochs, to observe the loss curve's further progression. However, as the loss plateaued at 10\textsuperscript{-5}, the model trained for 30 epochs was selected.

% The training loss plot vs the number of epochs is shown in Fig.~\ref{fig:training-loss-30}.
% The model was also trained without early stopping for 500 epochs, as shown in Fig.~\ref{fig:training-loss-500}, to observe the loss curve's further progression
% \begin{figure}[tb]
%      \centering
%      \begin{subfigure}[b]{0.45\textwidth}
%          \centering
%          \includegraphics[width=\textwidth]{images/Training-loss-500-3.png}
%          \caption{Without early stopping}
%          \label{fig:training-loss-500}
%      \end{subfigure}
%      \hfill
%      \begin{subfigure}[b]{0.495\textwidth}
%          \centering
%          \includegraphics[width=\textwidth]{images/training-loss-30-3.png}
%          \caption{With early stopping}
%          \label{fig:training-loss-30}
%      \end{subfigure}
%         \caption{Plot of mean squared error training loss across epochs}
%         \label{fig:MSE-epochs}
% \end{figure}

\subsubsection{Employing the trained model for grid removal}
During runtime, input ECG images are segmented and given to the trained DnCNN-based model. Each image is split into R, G, and B channels and split into overlapping 30$\times$30 patches to avoid boundary artifacts when re-stitched together \cite{xu2022research}. These patches are processed through the denoising CNN, yielding grid-removed patches that are stitched together using the Exponential Distance Weighted method by Wu et al. \cite{wu2021exponential}. This method applies distance-based weighting to the overlapping areas, ensuring smoother reconstruction. The result is a seamlessly reconstructed, clean ECG image post-grid removal.

\subsection{Region of interest detection}
The image, after text and grid removal, undergoes region of interest (ROI) detection to segment the leads. As standard ECG prints contain multiple rows of data, we split them into ECG strips to aid in mask retrieval and time-series data conversion. For the current study, we employed a histogram-based method for ROI detection to estimate the boundary regions between the ECG strips, aiming to identify the segments containing ECG time-series. We used the fact that rows of pixels with ECG segments typically have lower pixel intensities compared to the regions between segments, in dark foreground (black/grey) images.

\begin{figure}[tb]
     \centering
     \begin{subfigure}[b]{0.48\textwidth}
         \centering
         \includegraphics[width=\textwidth]{images/00006_lr.dat-grid-removed.png}
         \caption{Before region of interest detection}
         \label{fig:before-roi}
     \end{subfigure}
     \hfill
     \begin{subfigure}[b]{0.48\textwidth}
         \centering
         \includegraphics[width=\textwidth]{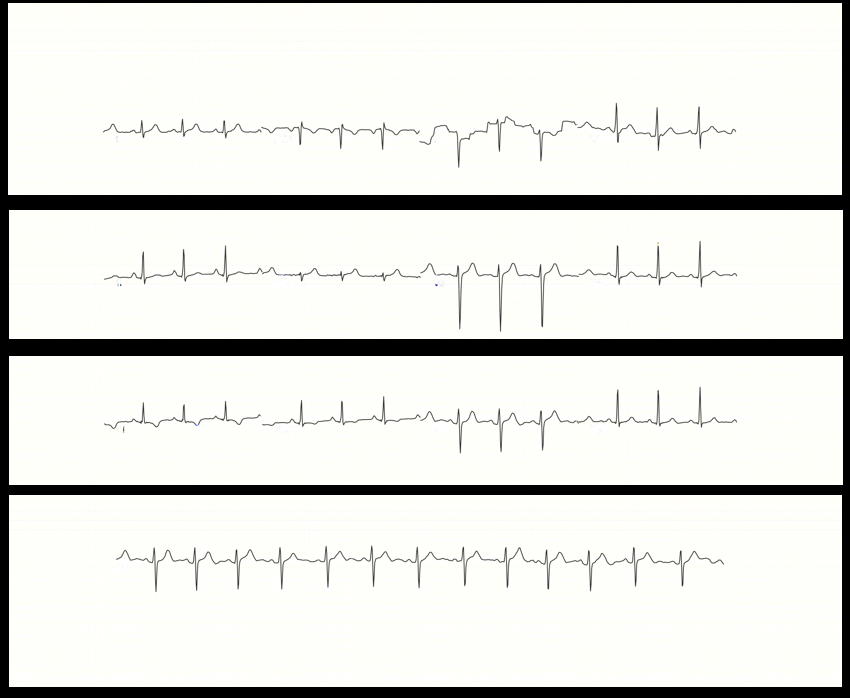}
         \caption{Separated rows of ECG}
         \label{fig:After-roi}
     \end{subfigure}
        \caption{Synthetic paper ECG image strips separated after region of interest detection}
        \label{fig:roi}
\end{figure}

Let $I(x, y)$ denote the pixel intensity value at coordinates $(x, y)$ of the image, where $x$ is the row index and $y$ is the column index. We compute the mean pixel intensity $\bar{I}(x) = \frac{1}{W} \sum_{y=1}^{W} I(x, y)$ row-wise by averaging intensities along each row, with $W$ being the image's width. Then, we plot pixel intensity vs row number, which typically shows global minima at ECG segments and global maxima between them. To smooth the pixel intensity curve and emphasize the global minima corresponding to ECG segments, we applied a non-causal moving average filter. The filtered pixel intensity curve is computed as:
\begin{equation}
\hat{I}(x) = \frac{1}{L} \sum_{k=-m}^{m} \bar{I}(x+k)
\end{equation}
where $L$ is the filter order ($L=11$ in the later reported results), and $m = (L-1)/2$. The separation regions between ECG segments are estimated as the rows between global minima. We then draw bounding boxes around the ECG signals using the estimated rows. These bounding boxes encapsulate the regions of interest containing individual ECG strips, which are then separated row-wise for subsequent processing and time-series data conversion. An example of the ROI detection results is shown in Fig.~\ref{fig:roi}.

More advanced ROI detection methods, like variants of the well-known You Only Look Once (YOLO) model \cite{Redmon_2016_CVPR}, can also be adapted for ECG applications. A pretrained YOLOv7 model is provided in ECG-Image-Kit for ROI detection.

\subsection{ECG time-series extraction}
The final step in the digitization pipeline is to convert segmented and denoised ECG segments into time-series data. In this stage, we use the fact that the ECG signal is a function (in the mathematical term), implying that when ECG segments are horizontally aligned (as a result of rotation compensation described in Section~\ref{sec:rotation_compensation}), there is exactly one corresponding ECG point per vertical column of the image segment. Thus, time-series recovery can be framed as identifying the most likely pixel per column. Furthermore, since the ECG waveform is a continuous time-series that adheres to the Nyquist frequency, it is continuous in time. Consequently, depending on the temporal and amplitude resolutions, the pixels most likely representing the ECG in adjacent columns are located near each other. In essence, the problem is a search for the most likely set of adjacent pixels forming the ECG curve, which can be done by a Viterbi search or similar methods \cite{Fortune2022}. In ECG-Image-Kit, several functions are provided for this stage. A simple method involves applying local smoothing filter followed by a column-wise peak search.

Here, we introduce a more detailed method using connected component analysis (CCA) on binarized segmented ECG images. CCA identifies contiguous regions, known as connected components, in a binary image based on a predefined connectivity criterion between pixels \cite{ganesh2021combining}. These connected components are labeled to uniquely identify each region for subsequent processing. The segmented ECG strips are given to CCA. The pixels within a connected component region receive unique labels through connected component labeling. Converting the ECG mask to time-series data can be complicated by discontinuities in the mask, which may arise from grid removal algorithms or residual artifacts in the image. CCA operates by scanning an image pixel-by-pixel and identifying regions with identical pixel intensities. For CCA, we first convert the image from RGB to grayscale and then apply a binary threshold. The binary threshold is implemented using a 3$\times$3 thresholding filter.

Next, we perform connected component labeling, fusing connected components within a specified distance threshold. Thresholds are set for the height, width, and area of the connected components. Components smaller than these thresholds are discarded, helping to eliminate stray pixels, character residuals and artifacts. Larger segments representing the background are also removed using a lower bound threshold. The distance threshold for fusing nearby components is determined using the Hausdorff distance~\cite{zhao2005new}, calculated using only the outer pixels of the components for computational efficiency. A connectivity of 4 for connected component labeling was found as optimal, considering the thickness of the ECG segments and the selected DPI of 200. This results in a mask for time-series extraction.

\begin{figure}[tb]
     \centering
         \includegraphics[width=.7\textwidth]{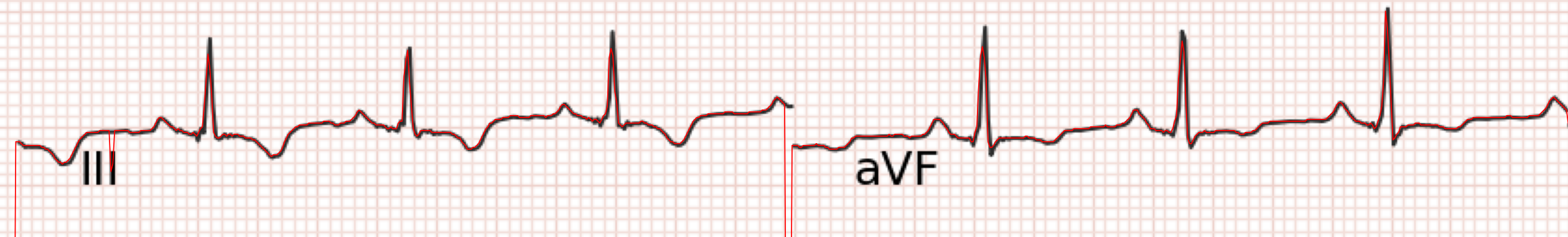}
         \caption{Obtained time-series ECG signal plotted in red superimposed on the synthetic paper ECG strip}
         \label{fig:final}
\end{figure}

After creating the connected mask, it undergoes blurring with a 1$\times$3 rectangular filter to capture neighborhood information along the horizontal (time) axis. This filter, focusing on filtering along the time axis, gathers contextual information for each pixel. The blurring filter's neighborhood information is crucial for determining whether a given pixel is part of the ECG signal, an assessment that can be framed as a maximum likelihood problem. We can gauge the likelihood of a pixel being part of the ECG signal based on its neighboring pixels' involvement in the signal. To extract the ECG signal pixels, we examine each column of the blurred mask, denoted by $B(x, y)$, and identify the pixel with the lowest intensity using:
\begin{equation}
    \hat{y}(x) = \arg\min_y B(x, y)
\end{equation}
where $\hat{y}(x)$ represents the pixel index with the lowest intensity in column $x$. This is based on the assumption of a dark foreground signal, where ECG pixels are generally darker than the background, making the pixel with the lowest intensity the most likely candidate to be part of the ECG signal. Sample ECG time-series overlaid on the ECG image segments are shown in Fig.~\ref{fig:final}.

\subsection{Conversion into physical units}
By this stage, we have a vector matching the ECG segment's pixel width, where each entry indicates the most likely vertical index of the ECG wave. To translate pixel indices to volts and seconds, we need the physical units per pixel. For accurately scanned images, this data can be derived from Section~\ref{sec:ecg_image_to_signal_resolution} or directly from the estimated ECG fine or coarse grid sizes. For the latter, considering each coarse grid equates to 0.5\,mV in amplitude, we use this amplitude scaling factor:
\begin{equation}
\text{Scaling Factor} = \frac{{0.5 \, \text{mV}}}{{\text{Coarse grid size in pixels}}}    
\end{equation}
Applying this scaling factor to the pixel index translates the data into a time-series representation in millivolts for the ECG image. As discussed in Section~\ref{sec:ecg_image_to_signal_resolution}, the sampling frequency of the extracted time-series is obtained from \eqref{eq:image_fs} (or can be inferred from the coarse or fine grid sizes in pixels). Importantly, this sampling frequency is different from the original time-series sampling frequency $f_s$. Hence, if required for waveform comparisons, the extracted signal should be resampled to $f_s$.

As a time-series, the extracted ECG waveforms may be further filtered using the state-of-the-art ECG filtering algorithms.

\subsection{Results}
We present the results of the ECG digitization pipeline in terms of quantitative signal quality metrics and ECG-based measurements.

\subsubsection{ECG time-series recovery performance}
The digitization pipeline, trained on synthetic paper ECG images from the PTB diagnostic ECG database \cite{bousseljot1995nutzung,Goldberger2000}, was evaluated on synthetic ECG images from a distinct dataset, the QT database \cite{laguna1997database,PhysioNetQT}. This dataset consisted of 1000 images, ensuring no overlap between the training and evaluation datasets. The evaluation set images were generated at a 200 DPI resolution and dimensions of 2200$\times$1700 pixels.

We employed both the standard signal-to-noise ratio (SNR) and an ad hoc SNR metric to assess the algorithm's performance in retrieving the ECG time-series. The standard SNR definition we used is:
\begin{equation}
    \text{SNR} = \frac{\text{\texttt{mean}}_k(x_k^2)}{\text{\texttt{mean}}_k[(x_k - \hat{x}_k)^2]}
\label{eq:snr_mean_based}
\end{equation}
which weights all sample points similarly. However, the standard SNR may not necessarily be the best metric for this problem from a practical perspective. In fact, one of the common issues in ECG digitization is that discrete mis-detected pixels may result in occasional spikes in the extracted time-series. While these spikes significantly impact the standard SNR metric, they are not necessarily the most detrimental for classification and human-based diagnosis applications. In fact, machine learning models, combined with appropriate filters, can remove spike noises. Human annotators are also adept at detecting unwanted spikes through visual inspection. Therefore, we may seek an SNR metric that is less susceptible to occasional spikes. Based on this observation, we propose the following modified SNR metric:
\begin{equation}
    \text{SNR}_\text{med} = \frac{\text{\texttt{mean}}_k(x_k^2)}{\text{\texttt{median}}_k[(x_k - \hat{x}_k)^2]}
\label{eq:snr_median_based}
\end{equation}
Accordingly, in $\text{SNR}_\text{med}$, instead of using the average of the noise power, we use the median of the noise power, which is more robust to occasional outliers.

The standard SNR and the median-based SNR were both calculated per record. For this purpose, all time series extracted from the digitization pipeline were resampled to their original sampling frequency (250\,Hz) and sample-wise aligned using the peak of their cross-correlation functions. Functions such as \texttt{finddelay} in MATLAB and \texttt{scipy.signal.correlation\_lags} in Python can be used for this alignment. This ensures that the algorithms are not disadvantaged or underrated due to missing only a few pixels at the beginning or end of the ECG segments.

\begin{figure}[tb]
     \centering
     \begin{subfigure}[b]{0.45\textwidth}
         \centering
         \includegraphics[width=\textwidth]{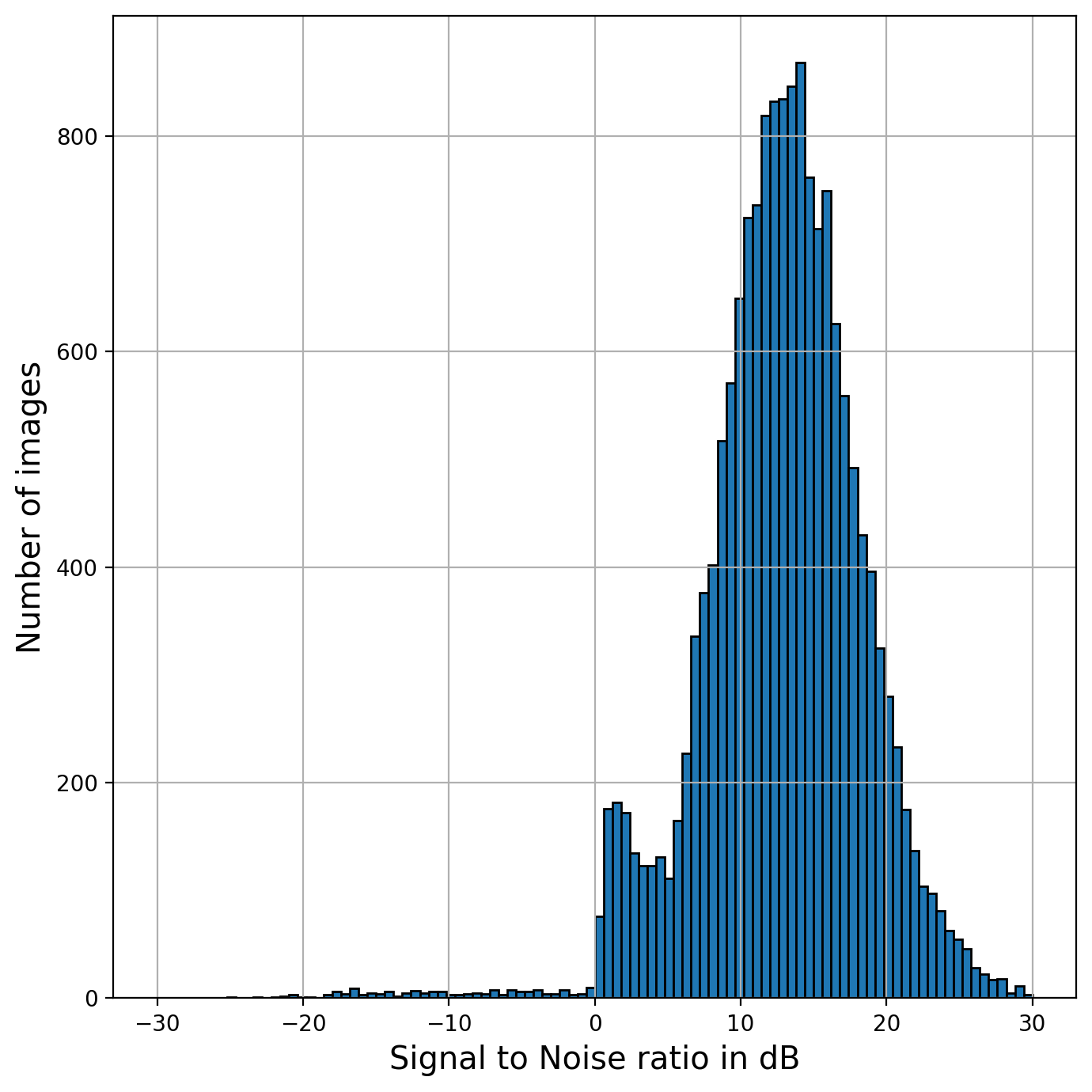}
         \caption{SNR}
         \label{fig:histogram-SNR-standard}
     \end{subfigure}
     \hfill
     \begin{subfigure}[b]{0.45\textwidth}
         \centering
         \includegraphics[width=\textwidth]{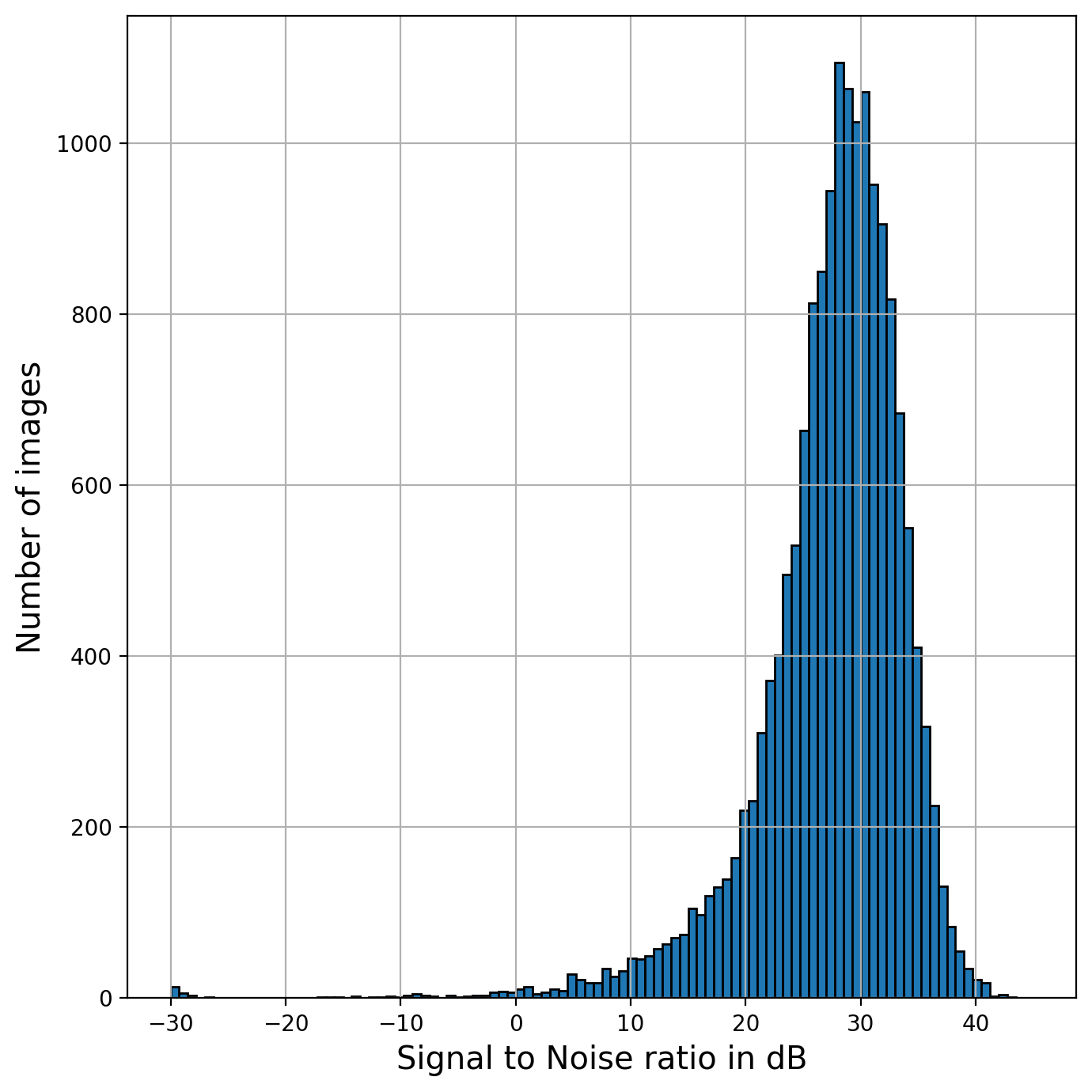}
         \caption{SNR\textsubscript{med}}
         \label{fig:histogram-SNR-median-noise}
     \end{subfigure}
        \caption{Histograms of the standard SNR metric (a) and the modified SNR using the median noise power (b) generated for the digitization of 8,505 synthetically generated images using real ECG samples from the QT database. The average and standard deviation of the standard SNR are 11.88\,dB and 8.91\,dB, respectively. The average and standard deviation of SNR\textsubscript{med} are 26.54\,dB and 10.11\,dB, respectively. Accordingly, the significantly higher values of SNR\textsubscript{med} over the conventional SNR metric confirms that spike noises are the major issue in the ECG extraction pipeline.}
        \label{fig:histogram-SNR}
\end{figure}

In Fig.~\ref{fig:histogram-SNR}, the histogram of SNR values for the 1000 evaluation images is shown. Table~\ref{table:evaluation_results_snr} details the evaluation results. The synthetic paper ECG dataset of 1000 images produced an average SNR of 11.88\,dB with a standard deviation of 8.91\,dB. The mean square error (MSE) was calculated as an additional evaluation metric, resulting in a mean 55.0\,$\mu$V and a standard deviation of 0.52\,mV for our synthetic paper ECG dataset. The mean and standard deviation of the modified SNR metric SNR\textsubscript{med} were 26.54\,dB and 10.11\,dB, respectively. Accordingly, the significant superiority of SNR\textsubscript{med} over the conventional SNR metric confirms that spike noises are the major issue in the ECG extraction pipeline. These results demonstrate the algorithm's ability to recover the ECG time series.

\begin{table}[tb]
  \centering
  \caption{Summary of the evaluation results using deep learning-based digitization algorithm}
  \label{table:evaluation_results_snr}
  \begin{tabular}{lcc}
    \toprule
    \textbf{Metric} & \textbf{Value} \\
    \midrule
    Average Signal to Noise Ratio (SNR) & 11.88\,dB \\
    Median based Average Signal to Noise Ratio (SNR) & 26.54 dB \\
    Standard Deviation of SNR & 8.91\,dB \\
    Mean Square Error (MSE) & 0.039\,mV \\
    Peak SNR & 31.26\,dB \\
    \bottomrule
  \end{tabular}
\end{table}
 
\subsubsection{Clinical parameter preservation}
\begin{figure}[tb]
     \centering
     \begin{subfigure}[b]{0.32\textwidth}
     \centering     \includegraphics[width=\columnwidth]{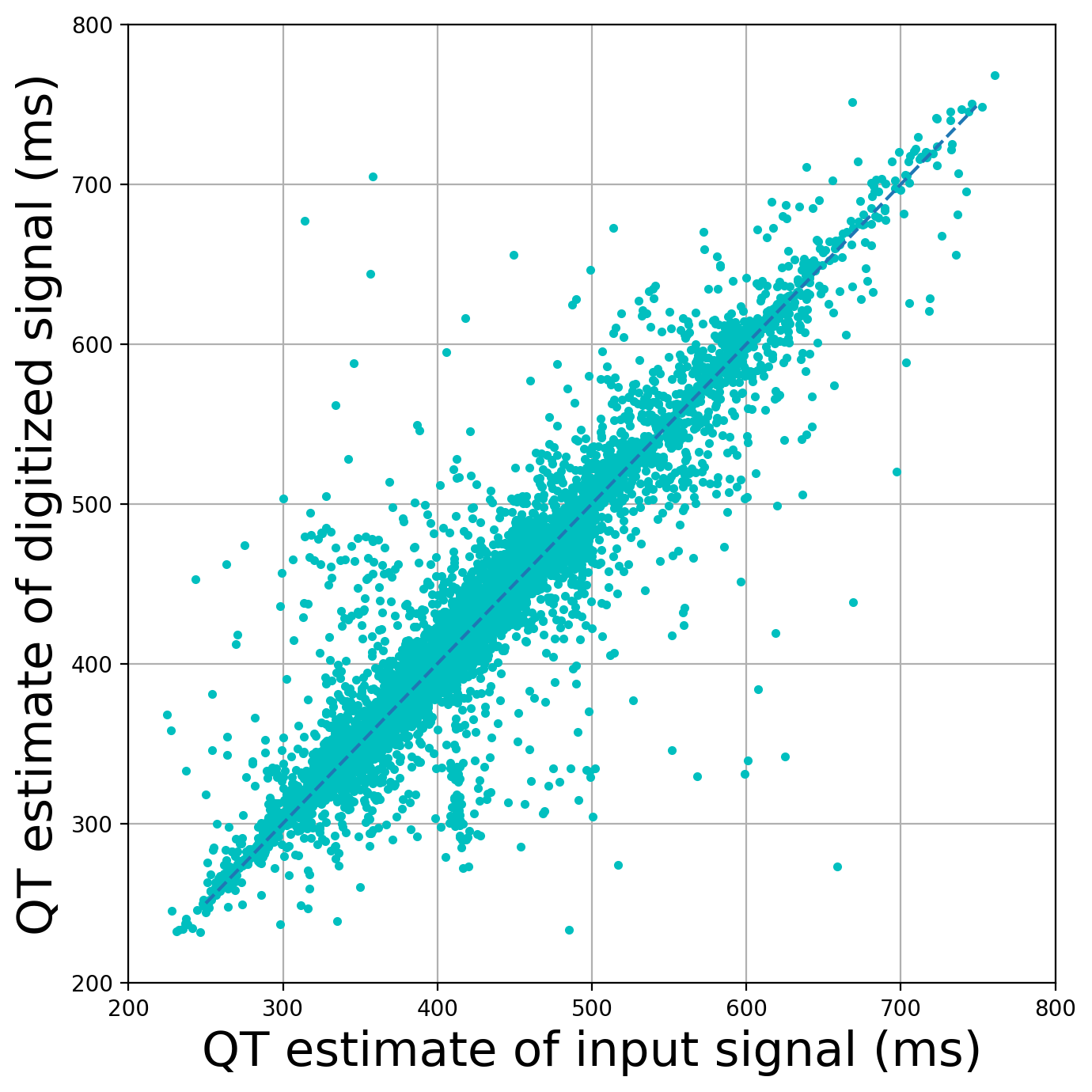}
         \caption{QT interval}
         \label{fig:QT-comparison}
     \end{subfigure}
     \hfill
     \begin{subfigure}[b]{0.32\textwidth}
     \centering     \includegraphics[width=\columnwidth]{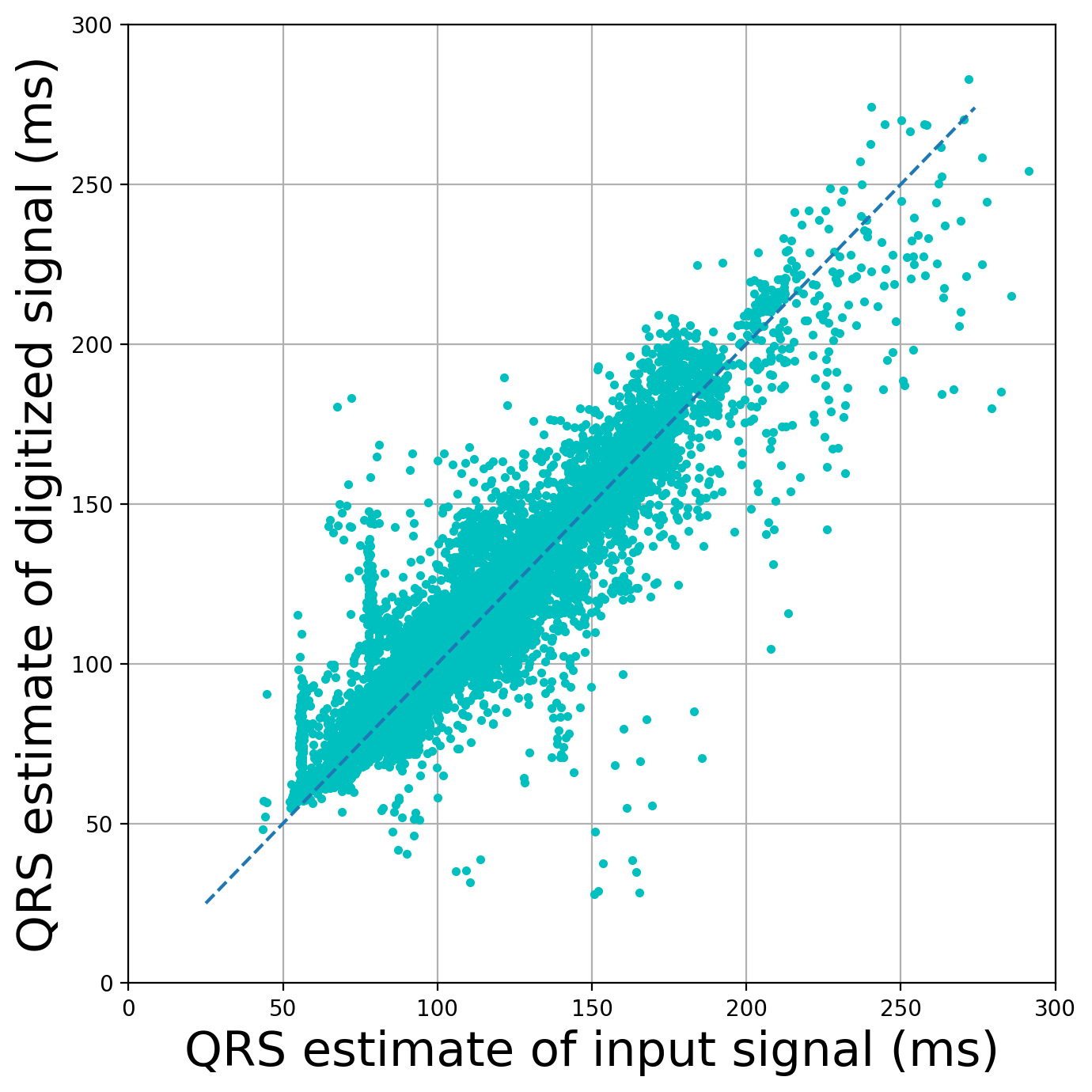}
         \caption{QRS width}
         \label{fig:QRS-comparison}
     \end{subfigure}
     \begin{subfigure}[b]{0.32\textwidth}
     \centering     \includegraphics[width=\columnwidth]{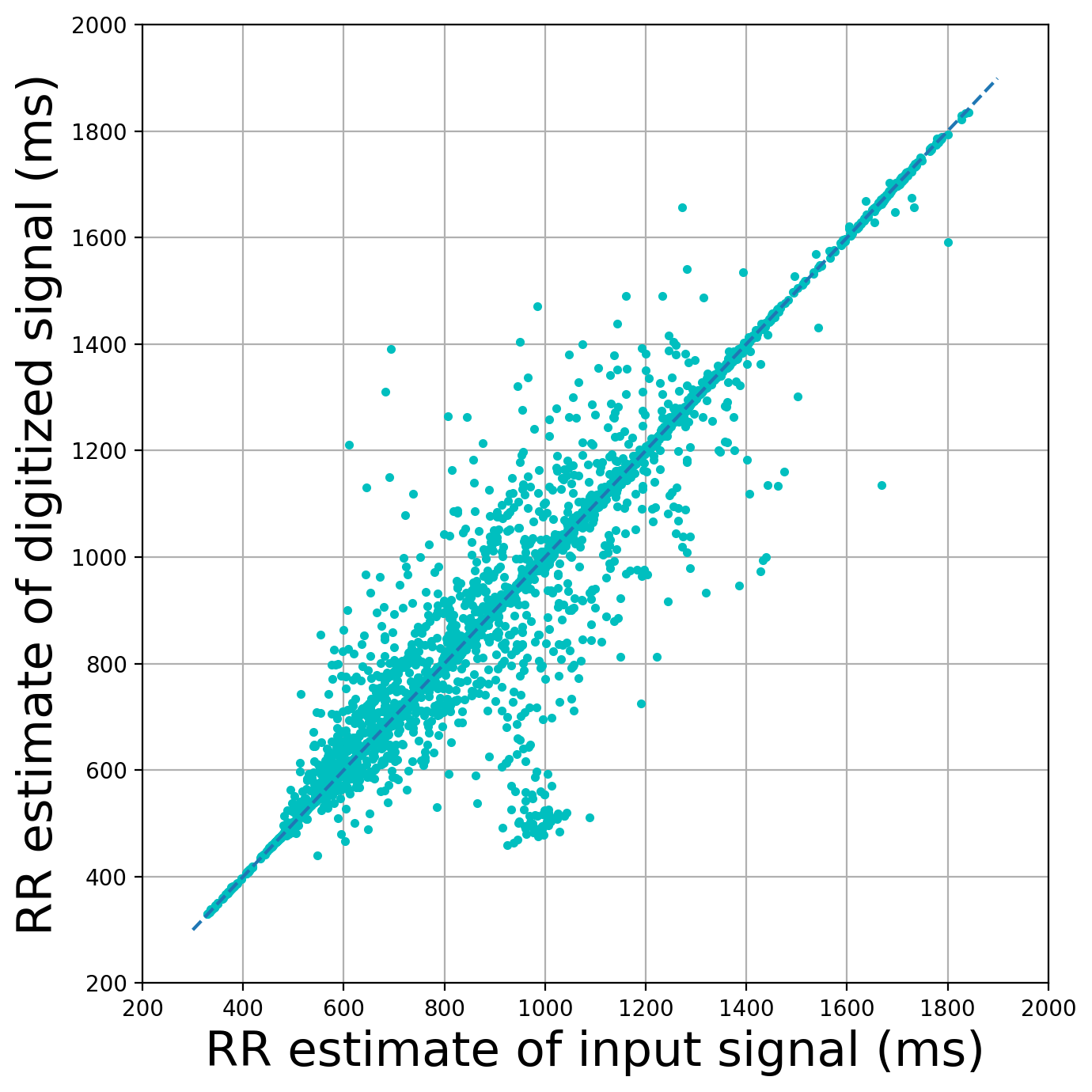}
         \caption{RR interval}
         \label{fig:RR-comparison}
     \end{subfigure}     
        \caption{Comparison of estimates of various clinical parameters extracted from time-series post ECG digitization vs the reference measurements from the original ECG time-series}
        \label{fig:clinical-params}
\end{figure}

While SNR is a standard metric of signal quality, the evaluation of the extracted ECG time-series should also be assessed in terms of the accuracy in extracting clinical biomarkers such as the RR interval (or its reciprocal, the heart rate), QRS width, QT interval, etc. We performed this evaluation on a dataset of 10,000 synthetically generated ECG images by the ECG-Image-Kit at a 200\,dpi resolution and of size 2200$\times$1700 pixels. The images were generated from different 10-second segments of the QT database \cite{laguna1997database,PhysioNetQT}. The database contains 100 fifteen-minute two-lead ECG recordings. The synthetically generated images were digitized using the proposed digitization pipeline. The QRS widths, RR intervals, and QT intervals were measured from the original ECG time-series and from the ECG estimate recovered from the images, using the \texttt{peak\_det\_likelihood.m} R-peak detector from OSET \cite{OSET3.14}. The R-peaks were next given to \texttt{wavedet\_3D.m} from ECG-Kit \cite{Demski2016}, to extract the fiducial points of the ECG, including the QRS onset/offset and the T-wave offset (from time-series and post-extraction from images). The accuracy of the estimated QRS widths, RR and QT intervals were used to determine how well clinical parameters were preserved throughout the digitization pipeline \cite{dumitru2023data}. The QRS widths, RR and QT-intervals of the reference ECG time-series data and the ones obtained from the digitized time-series data were measured and compared, as shown in Fig.~\ref{fig:clinical-params}. It can be seen from the corresponding figures that the reference and extracted measurements are highly correlated. Visual inspection of the outliers (for both SNR metrics and clinical parameters) revealed that the underperforming cases corresponded to the following: 1) mis-detected R-peaks and erroneous fiducial-point extraction due to extremely irregular beat morphologies and rhythms, which impact the clinical parameters but is not a shortcoming of the digitization pipeline; 2) low-amplitude ECG resulting in extensive quantization noise throughout the digitization pipeline; and 3) residual noise from remaining text or grid residuals, resulting in occasional spike noises in the extracted ECG time-series.

\section{Conclusion}
In this work, we introduced ECG-Image-Kit, a novel tool for creating synthetic paper-like ECG images from time-series data, and assessed its utility in a case study for training a comprehensive ECG image digitization pipeline combining image processing and deep learning techniques. The generated synthetic images effectively mimic real paper ECGs with realistic distortions such as handwritten text, wrinkles, and perspective changes. Using SNR, a modified SNR definition based on median noise power, and mean square error (MSE) metrics, we demonstrated the effectiveness of our toolset in accurately digitizing ECG images, evidenced by high SNR values (low MSE) and the close resemblance of synthetic images to original time-series data.

This research addresses the scarcity of real patient ECG data due to privacy and regulatory constraints. By generating synthetic ECG images, we provide a means to develop and test ECG analysis algorithms while ensuring privacy compliance. Our approach offers a diverse, controlled dataset that facilitates rigorous testing and enhancement of digitization techniques. This synthetic dataset is invaluable for developing algorithms, augmenting training data for machine learning models, and advancing automated ECG diagnoses.

In the clinical parameter extraction results, the measurements obtained from the original ECG time series were compared with those made after the ECG digitization pipeline. While this approach objectively assessed the performance of the digitization algorithm, the results are not indicative of the actual clinical parameter measurements, as they were also impacted by inaccuracies in R-peak detection and fiducial-point extraction algorithms, which are independent from the digitization algorithm. In the future, we can evaluate the performances of clinical parameters against human annotations.

Future research could expand the synthetic dataset with more variations like electrode misplacements, noise patterns, and heart rate variability. Collaborating with medical experts to conduct a \textit{Turing test} could validate the synthetic data's realism. Employing measures like the Kappa value in these tests would assess the perceptual fidelity of the synthetic dataset, confirming its utility in clinical settings for tasks like algorithmic ECG annotation and training. Furthermore, incorporating advanced techniques like generative adversarial networks (GANs) could result in complementary models with even more realistic and varied ECG types.

Optimizing the denoising CNN model and integrating advanced deep learning models or techniques like RNNs or attention mechanisms could refine the digitization process's accuracy and efficiency. Applying the pipeline to large datasets and diverse clinical scenarios will offer insights into its effectiveness across various ECG recording environments. Moreover, including domain-specific knowledge, like cardiac anatomical information or waveform characteristics, could enhance the digitization precision and lead to more accurate recovery and interpretation of ECG data.

This research therefore lays the groundwork for high-accuracy, generalizable ECG digitization solutions using synthetic ECG data, a critical step toward advancing ECG analysis in low resourced settings and enhancing global patient care standards.

% In conclusion, the study presented in this paper lays the foundation for employing digital twins in ECG analysis, opening doors to future advancements in medical diagnostics and contributing to improved patient care.

% \bibliographystyle{abbrvnat}
\bibliographystyle{IEEEtran}
\bibliography{egbib}

\end{document}